%% file: main.tex
\documentclass[10pt,twocolumn,letterpaper]{article}

\usepackage[dvipsnames,table]{xcolor}
\usepackage[pagenumbers]{iccv} 
\usepackage{booktabs}
\usepackage{pifont}

\usepackage{algorithm}
\usepackage{algpseudocode}

\usepackage{amsmath,amsfonts,bm}
\usepackage{bbm}
\usepackage[pagebackref,breaklinks,colorlinks,allcolors=cvprblue]{hyperref}


\definecolor{tickgreen}{HTML}{4f772d}
\definecolor{crossred}{HTML}{bc4749}
\definecolor{cvprblue}{rgb}{0.21,0.49,0.74}

\title{Is Less More? Exploring Token Condensation as Training-free \\Test-time Adaptation}

\author{ Zixin Wang$^1$, Dong Gong$^2$, Sen Wang$^1$, Zi Huang$^1$, Yadan Luo$^1$\\
$^1$The University of Queensland, Australia\\
$^2$University of New South Wales, Australia\\
{\tt\small \{zixin.wang,sen.wang,helen.huang,y.luo\}@uq.edu.au dong.gong@unsw.edu.au}}

\begin{document}
\maketitle
\input{sec/0_abstract}    
\input{sec/1_intro}

\input{sec/2_related_work}
\input{sec/3_method}

\input{sec/4_experiments}
\input{sec/5_conclusion}
{
    \small
    \bibliographystyle{ieeenat_fullname}
    \bibliography{main}
}
\input{sec/x_supp}

\end{document}

%% file: sec/0_abstract.tex
\begin{abstract}
Contrastive Language-Image Pretraining (CLIP) excels at learning generalizable image representations but often falls short in zero-shot inference on certain downstream datasets.
Test-time adaptation (TTA) mitigates this issue by adjusting components like normalization layers or context prompts, yet it typically requires large batch sizes and extensive augmentations, leading to high computational costs. This raises a key question: \textbf{Can VLMs' performance drop in specific test cases be mitigated through efficient, training-free approaches?}
To explore the solution, we investigate token condensation (TC) techniques, originally designed to enhance vision transformer efficiency by refining token usage during inference. We observe that informative tokens improve visual-text alignment in VLMs like CLIP on unseen datasets. However, existing TC methods often fail to maintain in-distribution performance when reducing tokens, prompting us to ask: How can we transform TC into an effective ``free-lunch'' adaptation strategy for VLMs? To address this, we propose Token Condensation as Adaptation (TCA), a training-free adaptation method that takes a step beyond standard TC. Rather than passively discarding tokens, TCA condenses token representation by introducing reservoir-based domain anchor tokens for information-preserving token reduction and logits correction. 
TCA achieves up to a 21.4\% performance improvement over the strongest baseline on cross-dataset benchmark and the CIFAR-100-Corrupted dataset while reducing GFLOPs by 12.2\% to 48.9\%, with minimal hyperparameter dependency on both CLIP and SigLIP series. 

\end{abstract}

%% file: sec/1_intro.tex
\section{Introduction}
\label{sec:intro}

\label{Sec:intro}
Online test-time adaptation (TTA) \cite{DBLP:journals/corr/abs-2310-20199} has emerged as a promising strategy to handle distribution shifts encountered during inference \cite{DBLP:journals/corr/abs-2303-15361}. TTA dynamically fine-tunes pretrained models on unlabeled data batches, enhancing generalization by aligning intermediate-layer batch statistics \cite{DBLP:conf/iclr/Niu00WCZT23}, optimizing for first-order flatness in the loss landscape \cite{DBLP:conf/iclr/ForetKMN21}, promoting self-supervised consistency across augmentations \cite{DBLP:conf/nips/ZhangLF22}, or tracking model historical weights \cite{DBLP:conf/icml/LeeC24a}. Despite the success of traditional TTA methods, they often require computationally expensive tuning of the backbone's parameters. This challenge is further exacerbated in CLIP \cite{DBLP:conf/icml/RadfordKHRGASAM21}, SigLIP \cite{DBLP:conf/iccv/ZhaiM0B23}, and SigLIP v2 \cite{tschannen2025siglip}, which rely on visual-text similarity for zero-shot prediction. With vast parameter sets and the need for large batch sizes (\eg, 256) to stabilize adaptation \cite{DBLP:journals/corr/abs-2405-14977}, applying conventional TTA to VLMs becomes increasingly impractical.

To circumvent the need for full model tuning, test-time prompting (TPT) has been proposed as a more efficient alternative. By learning a small set of task-specific context prompts, TPT better aligns text features with visual representations, enabling lightweight adaptation. However, TPT primarily focuses on refining text inputs while largely overlooking the impact of visual distribution shifts. Besides, adapting to high-variance target images through prompts often relies on external source data \cite{DBLP:conf/nips/SamadhGHKNK023} or extensive data augmentation \cite{DBLP:conf/iccv/Feng0LKZ23} (\eg, 60$\times$ more AugMix or diffusion-based synthetic samples). In strict online TTA settings, where the batch size is constrained to one, this reliance on augmentation significantly inflates computational costs, leading to a 60$\times$ rise in GFLOPs compared to single-sample processing (\ie, $1108.61$ vs. $17.59$ GFLOPs). The need for gradient backpropagation during inference further increases the computation burden, making exiting TPT suboptimal for many resource-constrained applications. 

The trade-off between the increased computation of TTA and the efficiency demands of testing raises a key question: Can VLMs achieve TTA while maintaining testing efficiency?
In this paper, we explore training-free solutions to address this challenge. Given that many VLM visual encoders rely on Vision Transformers (ViTs), token pruning and merging accelerate inference by reducing redundancy while preserving essential tokens through adjusted token usage in the forward pass. 
However, efficiency-driven token adjustments often come at the cost of degraded \textit{in-distribution} performance (\eg, ImageNet-1K validation) in exchange for lower GFLOPs \cite{DBLP:conf/iclr/LiangGTS0X22}. In contrast, our analysis reveals that selectively condensing low-attentiveness tokens not only preserves performance but can even enhance it on certain \textit{unseen} datasets (\cref{tab:cd-exp}). This insight motivates our exploration of efficient, training-free TTA through token condensation.

To better understand which tokens contribute most to VLM adaptation, we conducted a preliminary analysis on token importance, investigating their role in visual-text alignment (\cref{fig:leave-one-out}). This analysis highlights two key types of tokens that benefit from condensation: (1) class-irrelevant background tokens, which may mislead the model by emphasizing non-essential regions that deviate from pretraining data distributions, and (2) class-ambiguous object tokens, such as animal fur or textures, which overlap across categories and disperse visual embeddings. However, simply adapting to individual test samples in isolation is suboptimal, as it fails to capture domain-level trends that evolve over time. While textual features encode valuable class-specific information, their embedding space differs from visual representations, limiting direct alignment.

\begin{figure*}[t]
    \centering
    \includegraphics[width=1\linewidth]{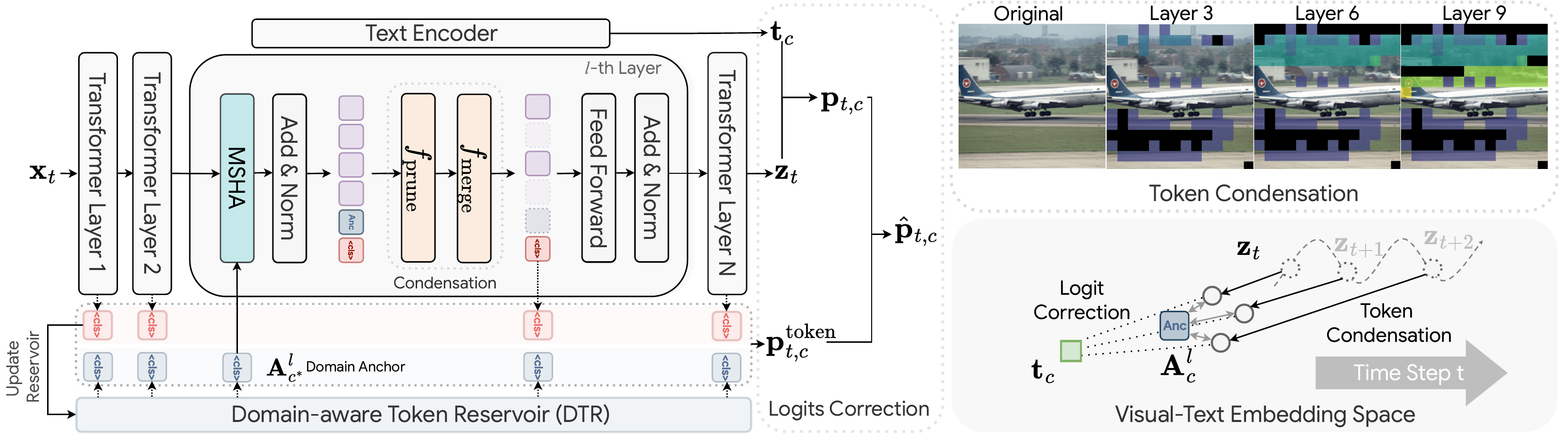}\vspace{-1.5ex}
    \caption{\textbf{Proposed Token Condensation as Adaptation (TCA).} Using CLIP as an example, to adapt visual embeddings to text embeddings during test-time, TCA utilizes a domain-aware token reservoir (DTR) to retain historical \texttt{<cls>} tokens with the lowest uncertainty as domain anchor tokens. These anchor tokens assist in (1) condensing tokens with low attentiveness scores \textit{(top-right)} and (2) acting as token-level classifiers to refine predictions through logits self-correction, moving visual embeddings $\mathbf{z}_t$ toward text embeddings $\mathbf{t}_c$.}
    \label{fig:overview}\vspace{-2ex}
\end{figure*}

To address these challenges, we introduce Token Condensation as Adaptation (TCA), a training-free adaptation method that dynamically refines token selection based on evolving domain knowledge. Instead of indiscriminately reducing tokens, TCA tracks and maintains domain-representative tokens over time, allowing the model to be adapted to new domains. As shown in \cref{fig:overview}, TCA stores domain-aware anchor tokens (\eg, \texttt{<cls>} tokens in CLIP or pooled vectors in SigLIP) in a reservoir (DTR), which then guides adaptation by refining class-specific representations. At each time step, TCA aligns token usage with domain trends by selectively adjusting token attentiveness based on both the current test sample and accumulated domain information in DTR. Additionally, TCA utilizes domain anchor tokens to refine model logits, improving visual-text alignment without modifying model parameters.

To our knowledge, this is the first work to explore token condensation as a form of test-time adaptation. Unlike previous approaches, TCA offers a lightweight, scalable, and training-free solution that generalizes across diverse data domains and can be readily extended to SigLIP and SigLIP v2. The extensive evaluations on the cross-dataset benchmark and the CIFAR-100-Corrupted dataset demonstrate that TCA consistently outperforms traditional TTA, prompting, and test-time prompting methods, achieving up to a 21.4\% improvement over the strongest baseline while reducing GFLOPs by 12.2\% to 48.9\%. 

%% file: sec/2_related_work.tex
\vspace{-1ex}
\section{Related Work}
\label{Sec:related_work}\vspace{-1ex}
\noindent\textbf{Online Test-time Adaptation.} To address performance degradation during test time, online test-time adaptation (TTA) has gained significant attention. Current TTA methods can be categorized into three main types \cite{DBLP:journals/corr/abs-2310-20199}: optimization-, data-, and model-based approaches. Optimization-based methods focus on model updates and optimization objectives \cite{DBLP:conf/nips/GoyalSRK22, DBLP:conf/wacv/MarsdenD024,DBLP:conf/nips/ZhangLF22}. A prominent example is Tent \cite{DBLP:conf/iclr/WangSLOD21}, which adapts Batch Normalization layers \cite{DBLP:conf/icml/IoffeS15} by entropy minimization. SAR \cite{DBLP:conf/iclr/Niu00WCZT23} extends this approach to Layer Normalization \cite{DBLP:journals/corr/BaKH16} and Group Normalization \cite{DBLP:conf/eccv/WuH18} with sharpness-aware minimization \cite{DBLP:conf/iclr/ForetKMN21}. Data-based methods include augmentations like selective \cite{DBLP:conf/cvpr/0013FGD22} and adversarial augmentation \cite{DBLP:conf/cvpr/TomarVBT23}, and memory banks \cite{DBLP:conf/nips/GongJKKSL22, DBLP:conf/cvpr/YuanX023,DBLP:conf/cvpr/0001WDE22}. Model-based approaches involve architectural modifications to enhance model adaptability during testing \cite{DBLP:conf/iclr/LiuYJZLGXZ24, DBLP:conf/nips/IwasawaM21, DBLP:conf/iclr/JangCC23, DBLP:conf/cvpr/WangZYZL23}. However, they typically depend on large batch sizes and augmentations, which introduce significant latency for online prediction. 

Recently, vision-language models like CLIP \cite{DBLP:conf/icml/RadfordKHRGASAM21} have excelled beyond fixed label sets, rendering traditional TTA methods less suitable \cite{DBLP:journals/corr/abs-2405-14977}. As a result, various online adaptation strategies have been proposed to improve zero-shot generalization. Test-time prompt tuning has emerged as a key approach in this context. TPT \cite{DBLP:conf/nips/ShuNHYGAX22} optimizes learnable prompts using data augmentations and soft entropy minimization, Diff-TPT \cite{DBLP:conf/iccv/Feng0LKZ23} enriches this with more diverse augmentations \cite{DBLP:conf/cvpr/RombachBLEO22}, while C-TPT \cite{DBLP:conf/iclr/YoonYTHLY24} focusing on model calibration. Other methods like VTE \cite{DBLP:journals/corr/abs-2405-14977} and DART \cite{DBLP:conf/aaai/Liu0PZ24} leverage prompt ensembles with DART further employing moving averages to boost performance. SwapPrompt \cite{DBLP:conf/nips/MaZG023} incorporates an EMA-updated target prompt. AdaPrompt \cite{DBLP:conf/aaai/Zhang0L24} utilizes a class-balanced memory bank to enhance adaptability. SCP \cite{wang2024towards} builds on TPT with a teacher-student framework to prevent semantic drift, while RLCF \cite{DBLP:conf/iclr/0006WZ024} incorporates reinforcement learning strategy \cite{DBLP:journals/ml/Williams92} to optimize the adaptation process. Beyond these, MTA \cite{DBLP:journals/corr/abs-2405-02266} introduces a new objective based on test-time augmentation to optimize visual features in the semantic space. TDA \cite{DBLP:journals/corr/abs-2403-18293} further improves CLIP's zero-shot ability by incorporating positive and negative caches with a training-free adapter. However, it relies on a large number of hyperparameters and is highly sensitive to them, while incurring significant computational costs during inference. In contrast, our approach strikes a better balance between computational efficiency and performance, outperforming both training-required and training-free methods.

\noindent\textbf{Token Condensation in Vision Transformers.} Vision transformers have achieved notable success in image recognition tasks, but their deployment is often limited by resource-constrained environments. To address this, various token condensation methods \cite{DBLP:conf/cvpr/MengLCLWJL22, DBLP:conf/nips/RaoZLLZH21, DBLP:conf/nips/RyooPADA21, DBLP:conf/aaai/XuZZSLDZXS22, DBLP:conf/eccv/ZongLSWQLL22, DBLP:conf/eccv/KongDMMNSSYRTQW22, DBLP:journals/corr/abs-2305-17328} have been proposed to reduce the computational overhead, primarily through two strategies: token pruning and token merging. Token pruning eliminates less informative tokens to save computation, as seen in methods like EViT \cite{DBLP:conf/iclr/LiangGTS0X22}, which retains tokens based on their attentiveness to the \texttt{<cls>} tokens. ATS \cite{DBLP:conf/eccv/FayyazKJSJSPG22} introduces input-dependent token pruning to adapt to variability across inputs. Token merging, on the other hand, seeks to combine similar tokens to reduce redundancy. For instance, ToME \cite{DBLP:conf/iclr/BolyaFDZFH23} uses bipartite soft matching to merge neighboring tokens that exhibit similarity. Hybrid approaches have also emerged, such as TPS \cite{DBLP:conf/cvpr/WeiYZTL23}, which prunes tokens and transfers information to retained ones using nearest-neighbor matching, and PruMerge \cite{DBLP:journals/corr/abs-2403-15388}, which prunes inattentive tokens using interquartile range and merges via k-nearest neighbors. While previous works have focused on enhancing efficiency within pure ViT models, our approach goes far beyond a mere adaptation of these methods, addressing multimodal distribution shifts in VLMs. This shift remains underexplored, particularly in how to use semantic guidance to prune irrelevant visual tokens that introduce ambiguity. By condensing these tokens, we effectively reduce such distribution shifts, enhancing test-time performance while simultaneously lowering computational costs (See \cref{tab:cd-exp}). 

%% file: sec/3_method.tex
\section{Token Condensation as Adaptation}\label{Sec:method}
Without loss of generality, we use CLIP as the representative model in the rest of the sections. We show our method can be effortlessly extended to other VLMs in \cref{sec:ablation}.

\noindent \textbf{Problem Set-up.} 
We begin by revisiting online test-time adaptation. For a given downstream task $\mathcal{D}_{\operatorname{tar}}$, the test data $\mathbf{x} = \{\mathbf{x}_t\}_{t=1}^T$ arrives sequentially at each time step $t$. The objective is to adapt the model on the fly to classify the incoming test samples into one of $C$ classes, each represented by a textual prompt like ``\texttt{a photo of a <classname>}''. CLIP embeds both visual and textual inputs into a shared space. The visual encoder $E_v$ extracts visual features $\mathbf{z}_t = E_v (\mathbf{V}_t) \in\mathbb{R}^{D}$ from image patches $\mathbf{V}_t = [\mathbf{v}_{\operatorname{cls}}, \mathbf{v}_1, \dots, \mathbf{v}_N]\in\mathbb{R}^{(N+1) \times D_{v}}$ of dimension $D_{v}$, where $\mathbf{v}_{\operatorname{cls}}$ is a CLIP-only \texttt{<cls>} token appended to $N$ patches. The text encoder $E_t$ generates class embeddings $\mathbf{T} = \{\mathbf{t}_c\}_{c=1}^C $, where each $\mathbf{t}_c\in\mathbb{R}^{D}$ corresponding to a class prompt. Classification is performed by computing the cosine similarity between the visual embedding $ \mathbf{z}_{t}$ and each class embedding $ \mathbf{t}_c$ with the probabilities calculated as:
\begin{equation}
   \mathbf{p}_{t,c}(\mathbf{z}_t, \mathbf{t}_c) = \frac{\exp\left( \operatorname{cos}(\mathbf{z}_t, \mathbf{t}_c) / \tau \right)}{\sum_{j=1}^{C} \exp\left( \operatorname{cos}(\mathbf{z}_t, \mathbf{t}_j) / \tau \right)},
\end{equation}
where $\tau$ denotes the temperature parameter controlling the sharpness of the output distribution.

For the visual encoder of ViT-based CLIP, given an $L$-layer ViT, the forward pass through the $l$-th Transformer block, where $l \in [1, 2, \dots, L]$, is formulated as:
\begin{equation}
    \mathbf{V}^{l+1} =  \hat{\mathbf{V}}^{l} + \operatorname{MLP}( \hat{\mathbf{V}}^{l}),
\end{equation}
\begin{equation}
    \hat{\mathbf{V}}^{l} = \mathbf{V}^{l} + \frac{1}{H}\sum_{h=1}^H\operatorname{Attention}(\mathbf{V}^l\mathbf{W}_Q^h, \mathbf{V}^l\mathbf{W}_K^h) \mathbf{V}^{l}\mathbf{W}_V^h, \notag
\end{equation}
where $\mathbf{V}^l \in \mathbb{R}^{(N+1) \times D_v}$ is the token embeddings at layer $l$. The matrices $\mathbf{W}^h_Q$, $\mathbf{W}^h_K$, $\mathbf{W}^h_V\in\mathbb{R}^{D_v\times D_v}$ are the linear projection matrices for the query, key, and value vectors in the $h$-th attention head, respectively, within the total number of attention heads $H$.

\subsection{Empirical Discussion}
\textbf{Pitfalls of TTA.} In CLIP, since the target domain $\mathcal{D}_{\operatorname{tar}}$ is unseen during pre-training, the alignment between visual embeddings $\mathbf{z}_t$ and the textual embeddings $\mathbf{T}$ may be suboptimal. Previous methods have attempted to address this by learning domain-specific prompts \citep{DBLP:conf/iclr/YoonYTHLY24} or replacing classifier weights with visual centroids \citep{DBLP:conf/nips/IwasawaM21} to move $\mathbf{T}$ closer to $\mathbf{z}_t$. However, the variability in CLIP's visual embeddings is often much \textit{higher} than in textual embeddings \citep{DBLP:conf/icml/RadfordKHRGASAM21}. At the patch level, individual tokens within the visual embeddings can drift and vary significantly \citep{DBLP:conf/icml/RadfordKHRGASAM21}. Thus, it becomes more urgent to derive methods that adjust $\mathbf{z}_t$ towards $\mathbf{T}$ for improved alignment.
\begin{figure}[t]
    \centering
    \begin{subfigure}[t]{0.48\linewidth}
        \centering
        \includegraphics[width=\linewidth]{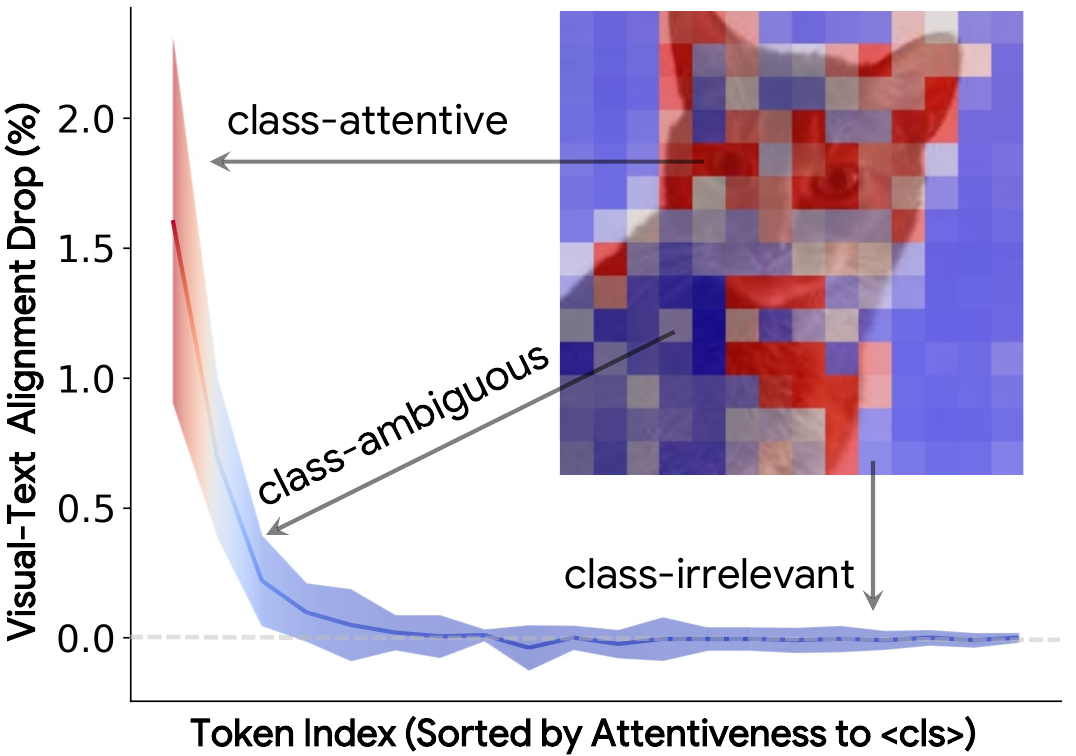}
        \caption{Impact of token removal on alignment. The warmer color indicates higher attentiveness.}
        \label{fig:leave-one-out}
    \end{subfigure}%
    \hfill
    \begin{subfigure}[t]{0.49\linewidth}
        \centering
        \includegraphics[width=\linewidth]{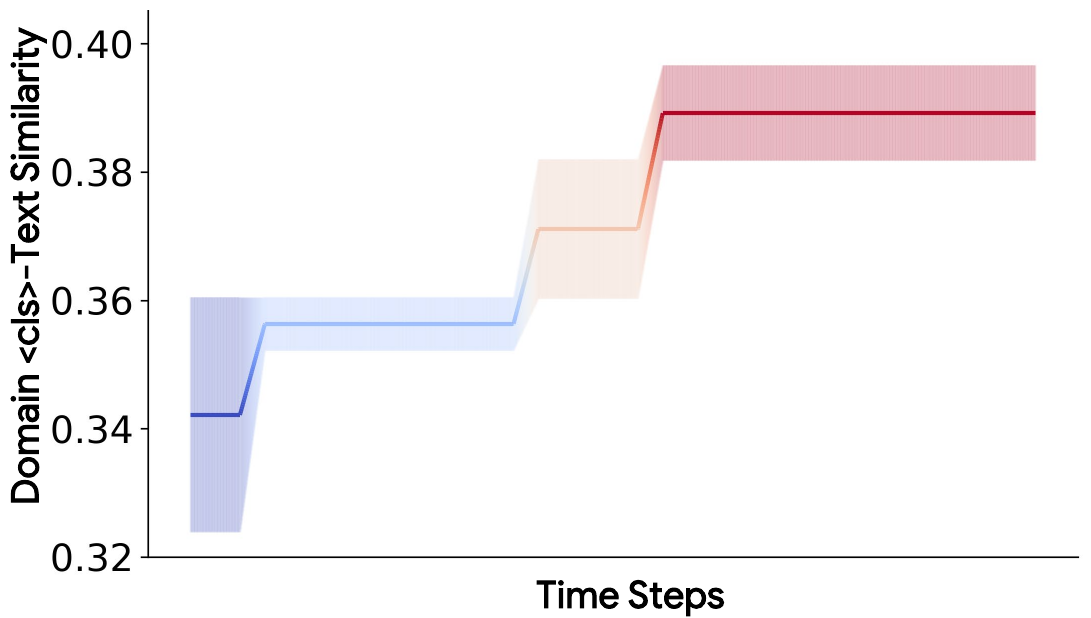}
        \caption{Alignment between the updated domain anchor token and text embedding over time.}
        \label{fig:anchor2text}
    \end{subfigure}\vspace{-1.5ex}
    \caption{Empirical studies of token influence and the strategy of caching domain anchor token (\ie, \texttt{<cls>} tokens in CLIP) to improve alignment.}\vspace{-2ex}
    \label{fig:enter-label}
\end{figure}

\noindent \textbf{How to Mine \texttt{<cls>} Tokens Aligned with Text?} The \texttt{<cls>} token in CLIP visual encoder is trained for broad concept alignment, often extending beyond target classes. Due to the mismatch between textual and visual token spaces, a key challenge is finding a more representative visual \texttt{<cls>} token that better aligns with text embeddings. To address this, we track visual \texttt{<cls>} tokens from test samples with the lowest entropy as \textbf{domain anchors}, updating them dynamically at each time step $t$. We then measure the alignment between these stored \texttt{<cls>} tokens and text embeddings $\mathbf{T}$ over time (\cref{fig:anchor2text}). Our results reveal a progressive correlation -- as the stored domain anchors consistently favor lower-entropy samples, their \texttt{<cls>} tokens increasingly align with text embeddings, effectively bridging the gap between visual and textual representations.

\subsection{Method Overview}
Building on our empirical findings, we propose Token Condensation as Adaptation (TCA), a training-free online adaptation strategy that enhances VLMs by filtering out tokens that contribute to visual-text misalignment, ensuring stable and efficient adaptation.

To leverage VLMs' existing knowledge, TCA introduces a domain-aware token reservoir (DTR) that retains representative anchor tokens for adaptation (\cref{sec.reservoir}). Note that in models like SigLIP \cite{tschannen2025siglip} and SigLIP v2 \cite{tschannen2025siglip}, which lack a \texttt{<cls>} token, we use the pooled feature vector to align with their holistic aggregation strategy. These domain anchors provide a stable reference for adaptation. Guided by these anchors, we perform cross-head token condensation between multi-head self-attention and feed-forward layers \citep{DBLP:conf/iclr/LiangGTS0X22}, selectively merging or discarding less informative tokens to ensure that only relevant, domain-consistent information is retained (\cref{sec:pruning}). Finally, stored domain anchor tokens aid in logits self-correction, refining model predictions based on accumulated domain knowledge (\cref{sec.logit_corr}). 



\subsection{Domain-aware Token Reservoir} \label{sec.reservoir}
To keep representative anchor tokens at domain level, we introduce domain-aware token reservoir $\mathfrak{R} = \{\mathfrak{R}_c\}_{c=1}^C$.
Notably, both token reduction and logits correction are building on the top of these saved tokens. In $\mathfrak{R}$, each buffer $\mathfrak{R}_c = \{(\mathbf{H}_c(\mathbf{z}_i; \mathbf{t}_c), \mathbf{A}_{i,c}^{\operatorname{cls}})\}_{i=1}^M$ is structured as a \textit{priority queue} that retains the top $M$ most reliable domain anchor tokens of target samples, which serve to implicitly distil semantic information from the corresponding text prompt $\mathbf{t}_c$ to guide the visual adaptation. These domain anchor tokens are crucial \textit{alignment proxies}: although the architectures of the text encoder $E_t$ and the visual encoder $E_v$ differ, the selected domain anchor tokens help determine which visual tokens best align with text features. The reliability of these domain anchor tokens is quantified by entropy scores, 
\begin{equation}
    \mathbf{H}_c(\mathbf{z}_t, \mathbf{t}_c) = -\mathbf{p}_{t,c}\left(\mathbf{z}_t, \mathbf{t}_c\right)\log\mathbf{p}_{t,c}\left(\mathbf{z}_t, \mathbf{t}_c\right),
\end{equation}
which act as keys to update the reservoir $\mathfrak{R}_c$. At each time step $t$, for each visual embedding $\mathbf{z}_t$, the corresponding anchor embeddings from all $L$ layers $\mathbf{A}_{t,c}^{\operatorname{cls}} =[\mathbf{v}_{\operatorname{cls}}^1,\ldots, \mathbf{v}_{\operatorname{cls}}^L]\in\mathbb{R}^{L\times D_v}$ will be stored in $\mathfrak{R}_c$ if $\operatorname{argmax}(\mathbf{p}_{t,c}) = c$, ensuring that only the most semantically consistent samples are retained:
\begin{equation} 
\mathfrak{R}_c \gets \operatorname{update}\left(\mathfrak{R}_c, \left(\mathbf{H}_c(\mathbf{z}_t, \mathbf{t}_c), \mathbf{A}_{t,c}^{\operatorname{cls}}\right)\right).
\end{equation}
If the priority queue $\mathfrak{R}_c$ has reached its capacity $M$, the sample with the highest entropy score is discarded and replaced with the new sample. Strategies for \textit{updating} the reservoir, such as first-in-first-out (FIFO) and similarity- or diversity-enforcing methods, are explored in \cref{sec:ablation}.

\subsection{Domain-aware Cross-head Token Reduction}\label{sec:pruning}
\begin{figure}
    \centering
    \includegraphics[width=0.9\linewidth]{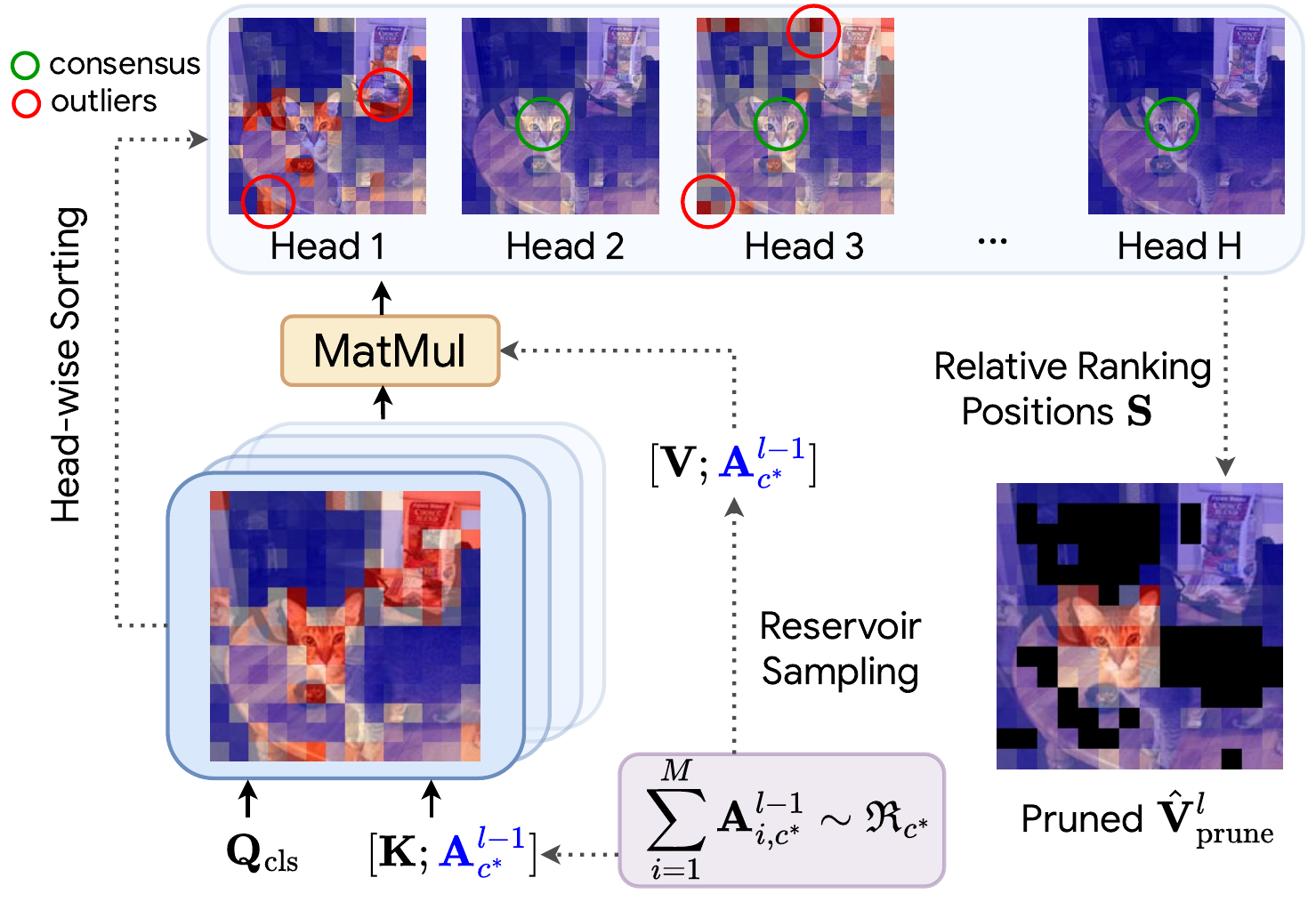}
    \caption{An overview of domain-aware cross-head token pruning.}\label{fig:pruning}\vspace{-1ex}
\end{figure}

Inspired by our preliminary studies as despite in \cref{fig:leave-one-out}, we introduce token condensation for training-free adaptation. Prior token reduction methods \citep{DBLP:conf/iclr/LiangGTS0X22} primarily discard patch tokens with lower averaged attention scores $\mathbf{S}\in\mathbb{R}^N$ relative to the \texttt{<cls>} token $\mathbf{v}_{\operatorname{cls}}^l$ across all attention heads $\mathbf{S}_i=\frac{1}{H}\sum_{h=1}^H\operatorname{Attention}(\mathbf{v}_{\operatorname{cls}}^l\mathbf{W}_Q^h, \mathbf{v}^l_i\mathbf{W}_K^h)$. However, this approach faces two limitations when applied in TTA tasks: \textbf{(1)} The \texttt{<cls>} token is \textit{universal} and may not be specifically aligned with the target class set. It may capture broad, unrelated semantics (\textit{e.g.,} ``cat food''), leading to the retention of irrelevant tokens that mislead the model into making incorrect predictions of the target class (\textit{e.g.,} ``cat''). \textbf{(2)} Averaging attention scores across all heads risks omitting important details, as each attention head tends to focus on distinct features (\textit{e.g.,} shape, color). Outliers in attention heads (highlighted by red circles in \cref{fig:pruning}) may disproportionately dominate the overall score, overshadowing valuable information by other heads. To overcome these limitations, we propose a \textbf{\textit{domain-aware cross-head token reduction}} that evaluates the token importance individually for each attention head with the consideration of saved domain anchor tokens and utilizes the averaged relative ranking positions to determine which tokens to prune and which to merge(see \cref{fig:pruning}). This approach reaches a more robust cross-head consensus and mitigates the impact of outliers. 

\noindent\textbf{Domain-aware Token Evaluation.} Firstly, we sample the domain anchor token for the $(l-1)$-th layer from domain-aware class buffer $\mathfrak{R}_{c^*}$, where $c^*$ is determined by the maximum cosine similarity between the current $\mathbf{v}^l_{\operatorname{cls}}$ token embedding and the average stored domain anchor tokens in $\mathfrak{R}_{c}$ for each class $c$. Here, the averaged domain anchor token is calculated by $\mathbf{A}_{c}^{l-1} = \frac{1}{M}\sum\nolimits_{i\in[M]} \mathbf{A}_{i, c}^{l-1}$. $c^* = \arg\max\nolimits_{c\in[C]} \operatorname{cos}(\mathbf{v}_{\operatorname{cls}}^l, \mathbf{A}_{i,c}^{l-1})$. Subsequently, we refine the attention map by concatenating $\mathbf{v}^l_{\operatorname{cls}}$ with the historical domain anchor token $\mathbf{A}_{c^*}^{l-1}$, 
\begin{equation}
    \operatorname{Attention}([\mathbf{v}_{\operatorname{cls}}^l;\mathbf{A}_{c^*}^{l-1}] \mathbf{W}^h_Q, [\mathbf{V}^l; \mathbf{A}_{c^*}^{l-1}]\mathbf{W}^h_K),\label{eq.domain-aware-eval}
\end{equation}
where $[\cdot;\cdot]$ indicates concatenation. This providing historical context that is \textit{better} aligned with target semantics.

\noindent\textbf{Cross-head Token Reduction. }To better evaluate the attentiveness of the tokens, we compute the token reduction score $\mathbf{S}^{\operatorname{head}}_i = \frac{1}{H}\sum_{h=1}^H\operatorname{rank}_h(i)$ for the $i$-th token, where $\text{rank}_h(i)$ gives the relative ranking position of token $i$ in head $h$ based on the attention score in \cref{eq.domain-aware-eval}. This ensures that tokens receiving consistently high attention across individual heads are retained, thereby achieving greater robustness to outliers in specific attention heads.

We find out the class-irrelevant and class-ambiguous tokens by the cross-head score $\mathbf{S}^{\operatorname{head}}_i$ and formulate token reduction in TCA as:
\begin{equation}
    \hat{\mathbf{V}}^{l} = f_{\operatorname{merge}}\circ f_{\operatorname{prune}}\left (\mathbf{V}^{l}; \mathfrak{R} \right),
\end{equation}
where $f_{\operatorname{prune}}(\cdot; \mathfrak{R}): \mathbb{R}^{N+1} \mapsto \mathbb{R}^{(\alpha\cdot R\cdot N) + 1}$ and $f_{\operatorname{merge}}(\cdot; \mathfrak{R}): \mathbb{R}^{(\alpha\cdot R\cdot N) + 1} \mapsto \mathbb{R}^{R \cdot N + 1}$ are pruning and merging respectively, responsible for reducing the number of tokens from $N + 1$ (including the \texttt{<cls>} token) to $R \cdot N + 1$, where $R$ is the fraction of tokens to be preserved. $\alpha$ controls the extent of token pruning. We first \textit{prune the class-irrelevant ones}:
\begin{equation}
\vspace{-1ex}
\hat{\mathbf{V}}_{\operatorname{prune}}^l \leftarrow \{\hat{\mathbf{v}}_i^l~|~\mathbf{S}^{\operatorname{head}}_i \leq \theta_{\operatorname{prune}}(\alpha, R), \forall i\in [N] \},\label{eq:token_pruning}
\end{equation}
where $\hat{\mathbf{V}}_{\operatorname{prune}}^l$ is the set of tokens retained after pruning at layer $l$. The pruning threshold $\theta_{\operatorname{prune}}(\alpha, R)$ allows only top-ranked $\alpha\cdot R \cdot N$ tokens are retained.

As depicted in \cref{fig:leave-one-out}, the \textit{class-ambiguous} tokens, although relevant to the target class, exhibit high uncertainty:
\begin{equation}
    \Phi = \{i~|~\theta_{\operatorname{merge}}(R) \leq \mathbf{S}^{\operatorname{head}}_i \leq \theta_{\operatorname{prune}}(\alpha, R), \forall i\},
\end{equation}
where $\theta_{\operatorname{merge}}(R)$ denotes thresholds for token selection during merging. The selected tokens $\mathbf{V}^l_{\Phi} = \{\mathbf{v}_i^l\}_{i\in\Phi}$ can introduce variance or noise into latent representation $\mathbf{z}_t$ and negatively impact the final classification decision. To address this, we propose a \textit{domain-aware token merging strategy} to consolidate these tokens into more representative ones. Here, rather than truncating neighbored token pairs with bipartite soft matching \citep{DBLP:conf/iclr/BolyaFDZFH23}, applying spectral clustering \citep{DBLP:conf/icml/BianchiGA20}, or graph pooling \citep{DBLP:conf/icml/WuC0L22}, we adopt a more efficient \textit{coreset selection} approach. Details of the coreset selection process and algorithm are provided in the supplementary material.
\subsection{Logits Self-correction} \label{sec.logit_corr}
To counter the shifts on the \textit{semantic} side after token condensation, we introduce a logits self-correction mechanism that leverages domain anchor tokens stored in $\mathfrak{R}$. In particular, the visual \texttt{<cls>} token of the current sample (if using CLIP) $\mathbf{V}_t^{\operatorname{cls}} \in \mathbb{R}^{L \times D_v}$ is compared with the stored domain anchor tokens $\mathcal{A} = \{\mathbf{A}_{i,c}^{\operatorname{cls}}\}_{i=1}^{M}$. 
The cosine similarity between these cross-layer tokens serves as a token-level classifier, which provides auxiliary information to adjust the predicted probability $\mathbf{p}_{t, c}$ from a \textit{visual} perspective:
 \begin{equation} \label{eq:logit_correction}
 \begin{aligned}
      &\tilde{\mathbf{p}}_{t,c} = \mathbf{p}_{t,c} + \lambda \mathbf{p}^{\operatorname{token}}_{t,c},
 \\ &\mathbf{p}^{\operatorname{token}}_{t,c} = \frac{1}{M} \sum_{i=1}^{M} \operatorname{cos}(\mathbf{V}_t^{\operatorname{cls}}, \mathbf{A}_{i,c}^{\operatorname{cls}})\cdot\mathbf{P} \cdot \mathbbm{1}_{c}, 
 \end{aligned}
 \end{equation} 
 where $\lambda$ is the logits correction weight. $\mathbbm{1}_c\in\mathbb{R}^{C}$ the one-hot vector for the $c$-the class. The layer-specific exponential scaling coefficients are denoted as $\mathbf{P}= [\exp(\frac{l}{\beta})]_{l=1}^L\in\mathbb{R}^L$, where $\beta$ controls the influence of different layers. We show that this correction temperature $\beta$ provides semantic interpretability, as further discussed in \cref{sec:ablation}. This self-correction mechanism ensures better alignment between the visual and semantic contexts, improving robustness in handling semantic shifts.

%% file: sec/4_experiments.tex
\section{Experiments}
\subsection{Experimental Setup}
\noindent\textbf{Datasets.} Following prior works, we evaluate our method on the cross-dataset (CD) benchmark, which measures model performance on unseen classes across 10 datasets: Aircraft \cite{DBLP:journals/corr/MajiRKBV13}, Caltech101 \cite{DBLP:journals/cviu/Fei-FeiFP07}, Cars \cite{DBLP:conf/iccvw/Krause0DF13}, DTD \cite{DBLP:conf/cvpr/CimpoiMKMV14}, EuroSAT \cite{DBLP:journals/staeors/HelberBDB19}, Flower102 \cite{DBLP:conf/icvgip/NilsbackZ08}, Food101 \cite{DBLP:conf/eccv/BossardGG14}, Pets \cite{DBLP:conf/cvpr/ParkhiVZJ12}, SUN397 \cite{DBLP:conf/cvpr/XiaoHEOT10}, and UCF101 \cite{DBLP:journals/corr/abs-1212-0402}. Additionally, we assess TCA's robustness to distribution shifts using CIFAR-100-Corrupted \cite{DBLP:conf/iclr/HendrycksD19} (CIFAR-100-C), which introduces varying corruption severities. Further details on additional experiments are provided in the supplementary material.

\noindent \textbf{Baselines.} We compare TCA with existing approaches across four categories: (1) \textit{Prompt-tuning methods} like CoOp \cite{DBLP:journals/ijcv/ZhouYLL22} and CoCoOp \cite{DBLP:conf/cvpr/ZhouYL022}, which require multi-epoch adaptation; (2) \textit{Conventional online test-time adaptation (TTA) methods} such as Tent \cite{DBLP:conf/iclr/WangSLOD21} and SAR \cite{DBLP:conf/iclr/Niu00WCZT23}. Tent updates batch normalization layers, while SAR further incorporates sharpness-aware minimization for reliable model updates. Following \cite{DBLP:journals/corr/abs-2405-14977}, we reran these experiments with adjusted batch sizes to align with our settings; (3) \textit{Test-time prompting methods}, including TPT \cite{DBLP:conf/nips/ShuNHYGAX22}, C-TPT \cite{DBLP:conf/iclr/YoonYTHLY24}, and Diff-TPT \cite{DBLP:conf/iccv/Feng0LKZ23}, as well as TTA methods for CLIP such as MTA \cite{DBLP:journals/corr/abs-2405-02266} and TDA \cite{DBLP:journals/corr/abs-2403-18293}; and (4) \textit{Token pruning and merging methods for ViTs}, such as EViT \cite{DBLP:conf/iclr/LiangGTS0X22}, ToMe \cite{DBLP:conf/iclr/BolyaFDZFH23}, and ATS \cite{DBLP:conf/eccv/FayyazKJSJSPG22}. As ATS is an adaptive token pruning method with no fixed budget, we constrain its computational cost by an upper bound to ensure fair comparison.

\noindent \textbf{Implementation Details.} For CLIP, we use its official prompts. The batch size is set to 1, without data augmentations, to mimic realistic deployment scenarios. All experiments are conducted using pre-trained CLIP models with ViT-B/16 and ViT-L/14 architectures as the visual backbone. We set $K$ to 2.
For SigLIP and SigLIP v2, we select models with ViT-B/16 as the backbone. Since the SigLIP series learns visual embeddings differently (\ie, via MAP rather than relying on the last layer's \texttt{<cls>} token), we adapt our method accordingly. Specifically, we use pooled attention weights to obtain the domain anchor and the token attentiveness score ($S_i^\text{head}$) for cross-head token reduction and logits self-correction. Unless otherwise stated, all ablation studies are conducted on CLIP with ViT-B/16. Notably, our method is training-free, enabling rapid adaptation with minimal hyperparameter dependency. All experiments are performed on a single NVIDIA RTX A6000 GPU.

\begin{table*}[th]
\caption{Results on the cross-dataset benchmark using CLIP ViT-B/16, including the number of learnable parameters (L-Param.) for learning-based TTA methods. $^*$ denotes the averaged GFLOPs across all datasets. The best performance (aug-free) is \textbf{bolded}. }\vspace{-1.5ex}
  \centering
  \resizebox{\linewidth}{!}{
    \begin{tabular}{l*{12}{c}lc}
      \toprule
      Method & \rotatebox{60}{Aug-free} & \rotatebox{60}{Aircraft} & \rotatebox{60}{Caltech101} & \rotatebox{60}{Cars} & \rotatebox{60}{DTD} & \rotatebox{60}{EuroSAT} & \rotatebox{60}{Flower102} & \rotatebox{60}{Food101} & \rotatebox{60}{Pets} & \rotatebox{60}{SUN397} & \rotatebox{60}{UCF101} & \rotatebox{60}{\textbf{Average}} & \rotatebox{60}{\textbf{GFLOPs}} & \rotatebox{60}{\textbf{L-Param.}}\\
      \midrule
      \midrule
      CLIP & \textcolor{tickgreen}{\ding{51}} & 23.22 & 93.55 & 66.11 & 45.04 & 50.42 & 66.99 & 82.86 & 86.92 & 65.63 & 65.16 & \cellcolor{pink!30}{64.59} & \cellcolor{blue!10}{17.59} & 0\\
      CoOp & \textcolor{crossred}{\ding{55}} & 18.47 & 93.70 & 64.51 & 41.92 & 46.39 & 68.71 & 85.30 & 89.14 & 64.15 & 66.55 & \cellcolor{pink!30}{63.88} & \cellcolor{blue!10}{17.59} & 2048\\
      CoCoOp & \textcolor{crossred}{\ding{55}} & 22.29 & 93.79 & 64.90 & 45.45 & 39.23 & 70.85 & 83.97 & 90.46 & 66.89 & 68.44 & \cellcolor{pink!30}{64.63} & \cellcolor{blue!10}{17.59} & 34,816\\
      \midrule
       Tent & \textcolor{tickgreen}{\ding{51}} & 8.97 & 93.39 & 62.69 & 39.78 & 20.85 & 61.23 & 83.70 & 87.76 & 65.30 & 66.93 & \cellcolor{pink!30}{59.06} & \cellcolor{blue!10}{17.59} & 40,960\\
      SAR & \textcolor{tickgreen}{\ding{51}} & 21.09 & 91.85 & 61.15 & 44.68 & 46.19 & 63.54 & 81.43 & 87.95 & 59.74 & 65.58 & \cellcolor{pink!30}{62.32} & \cellcolor{blue!10}{17.59} & 31,744\\
      TPT & \textcolor{crossred}{\ding{55}} & 24.78 & 94.16 & 66.87 & 47.75 & 42.44 & 68.98 & 84.67 & 87.79 & 65.50 & 68.04 & \cellcolor{pink!30}{65.10} & \cellcolor{blue!10}{1108.61} & 2048\\
      Diff-TPT & \textcolor{crossred}{\ding{55}} & 25.60 & 92.49 & 67.01 & 47.00 & 43.13 & 70.10 & 87.23 & 88.22 & 65.74 & 62.67 & \cellcolor{pink!30}{65.47} & \cellcolor{blue!10}{-} & -\\
      C-TPT & \textcolor{crossred}{\ding{55}} & 23.90 & 94.10 & 66.70 & 46.80 & 48.70 & 69.90 & 84.50 & 87.40 & 66.00 & 66.70 & \cellcolor{pink!30}{65.47} & \cellcolor{blue!10}{1108.61} & 2048\\
      MTA & \textcolor{crossred}{\ding{55}} & 25.32 & 94.21 & 68.47 & 45.90 & 45.36 & 68.06 & 85.00 & 88.24 & 66.67 & 68.11 & \cellcolor{pink!30}{65.53} & \cellcolor{blue!10}{-} & -\\

      TDA & \textcolor{tickgreen}{\ding{51}} & 23.91 & \textbf{94.24} & \textbf{67.28} & \textbf{47.40} & 58.00 & 71.42 & 86.14 & 88.63 & \textbf{67.62} & 70.66 & \cellcolor{pink!30}{67.53} & \cellcolor{blue!10}{17.59} & 0\\
      \midrule
      EViT$_{R=0.9}$ & \textcolor{tickgreen}{\ding{51}} & 24.12 & 92.25 & 64.57 & 45.09 & 48.41 & 70.24 & 84.99 & 88.96 & 64.58 & 68.46 & \cellcolor{pink!30}{65.17} & \cellcolor{blue!10}{15.41} & 0 \\
      ToME$_{R=0.9}$ & \textcolor{tickgreen}{\ding{51}} & 24.66 & 92.49 & 63.10 & 44.92 & 48.64 & 69.22 & 85.04 & 87.90 & 64.22 & 68.62 & \cellcolor{pink!30}{64.88} & \cellcolor{blue!10}{15.31} & 0 \\
      ATS$_{R=0.9}$ & \textcolor{tickgreen}{\ding{51}} & 22.86 & 92.21 & 57.90 & 40.96 & 40.62 & 67.52 & 80.16 & 85.34 & 61.53 & 67.22 & \cellcolor{pink!30}{61.63} & \cellcolor{blue!10}{11.15$^*$} & 0 \\
      EViT$_{R=0.7}$ & \textcolor{tickgreen}{\ding{51}} & 23.31 & 91.20 & 58.44 & 43.32 & 43.26 & 67.11 & 79.70 & 85.77 & 61.41 & 66.69 & \cellcolor{pink!30}{62.02} & \cellcolor{blue!10}{11.62} & 0 \\
      ToME$_{R=0.7}$ & \textcolor{tickgreen}{\ding{51}} & 22.26 & 90.79 & 55.48 & 42.32 & 40.12 & 64.11 & 79.36 & 84.19 & 60.66 & 63.97 & \cellcolor{pink!30}{60.33} & \cellcolor{blue!10}{11.45} & 0 \\
      ATS$_{R=0.7}$ & \textcolor{tickgreen}{\ding{51}} & 17.28 & 85.40 & 33.65 & 36.52 & 27.79 & 52.62 & 55.97 & 72.94 & 48.82 & 56.44 & \cellcolor{pink!30}{48.74} & \cellcolor{blue!10}{8.76$^*$} & 0 \\
      \midrule
      \textbf{TCA$_{R=0.9}$} & \textcolor{tickgreen}{\ding{51}} & \textbf{24.87} & 93.63 & 65.33 & 46.16 & \textbf{70.43} & \textbf{73.33} & \textbf{85.31} & \textbf{89.53} & 65.92 & \textbf{72.38} & \cellcolor{pink!30}{\textbf{68.69}} & \cellcolor{blue!10}{\(\mathbf{15.45}_{\mathbf{\textcolor{blue}{-12.2\%}}}\)} & 0 \\
      \textbf{TCA$_{R=0.7}$} & \textcolor{tickgreen}{\ding{51}} & 23.19 & 92.13 & 58.15 & 44.50 & 61.63 & 69.79 & 79.99 & 85.99 & 61.89 & 67.38 & \cellcolor{pink!30}{64.46} & \cellcolor{blue!10}{\(\mathbf{11.69}_{\mathbf{\textcolor{blue}{-33.5\%}}}\)} & 0 \\
      \bottomrule
    \end{tabular}}\vspace{-3ex}
  \label{tab:cd-exp}
\end{table*}
\vspace{-1ex}
\subsection{Main Results}
\cref{tab:cd-exp} presents the results for fine-grained cross-dataset benchmark using the ViT-B/16 architecture on CLIP. As observed in \cref{fig:leave-one-out}, the core idea behind TCA is that condensing inattentive tokens can effectively mitigate distribution shifts caused by visual-text misalignment. This concept is first validated by the improved performance of token pruning baselines over CLIP inference, where a condensed token set yields a $0.9\%$ increase in average accuracy when $R=0.9$. TCA further enhances its performance by dealing with visual-text misalignment, moving visual features toward historical domain anchor tokens from DTR. As a result, TCA achieves an average accuracy of 68.69\%, outperforming both train-required and training-free baselines without augmentation. Conventional TTA methods perform poorly on all datasets even with the requirement of fine-tuning a large amount of learnable parameters. In contrast, prompt-tuning methods, although requiring fewer learnable parameters, rely heavily on augmentation and struggle to effectively handle visual shifts. While TDA is a training-free method, it requires a large number of hyperparameters (a total of 10 for managing positive and negative caches) to achieve optimal performance. On the other hand, TCA uses significantly fewer hyperparameters and delivers a 1.72\% improvement in average accuracy over TDA, with approximately 12.2\% fewer GFLOPs. Further details on the impact of the visual backbone (ViT-L/14) and OOD benchmarks are provided in the supplementary material.
\begin{figure}[t]
    \centering
    \begin{subfigure}[t]{0.48\linewidth}
        \centering
        \includegraphics[width=\linewidth]{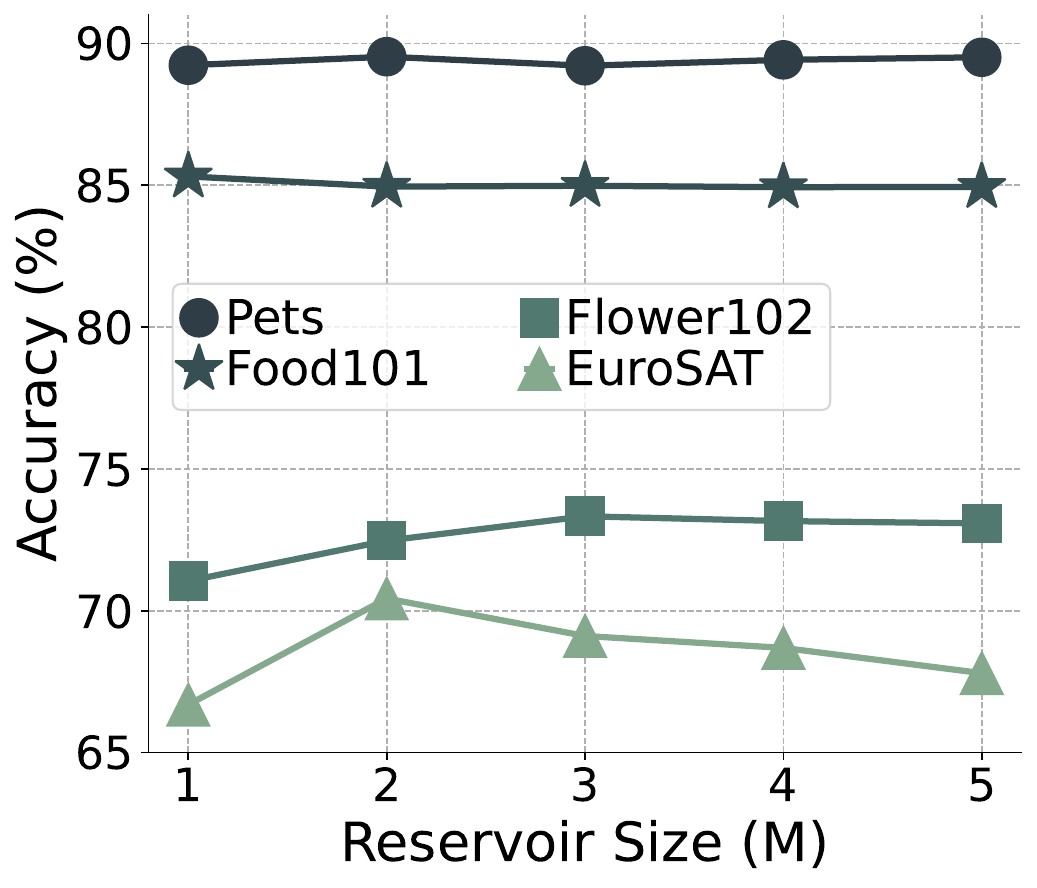}\vspace{-1ex}
        \caption{Impact of reservoir size.}
        \label{abl.cache_size}
    \end{subfigure}
    \begin{subfigure}[t]{0.48\linewidth}
        \centering
        \includegraphics[width=\linewidth]{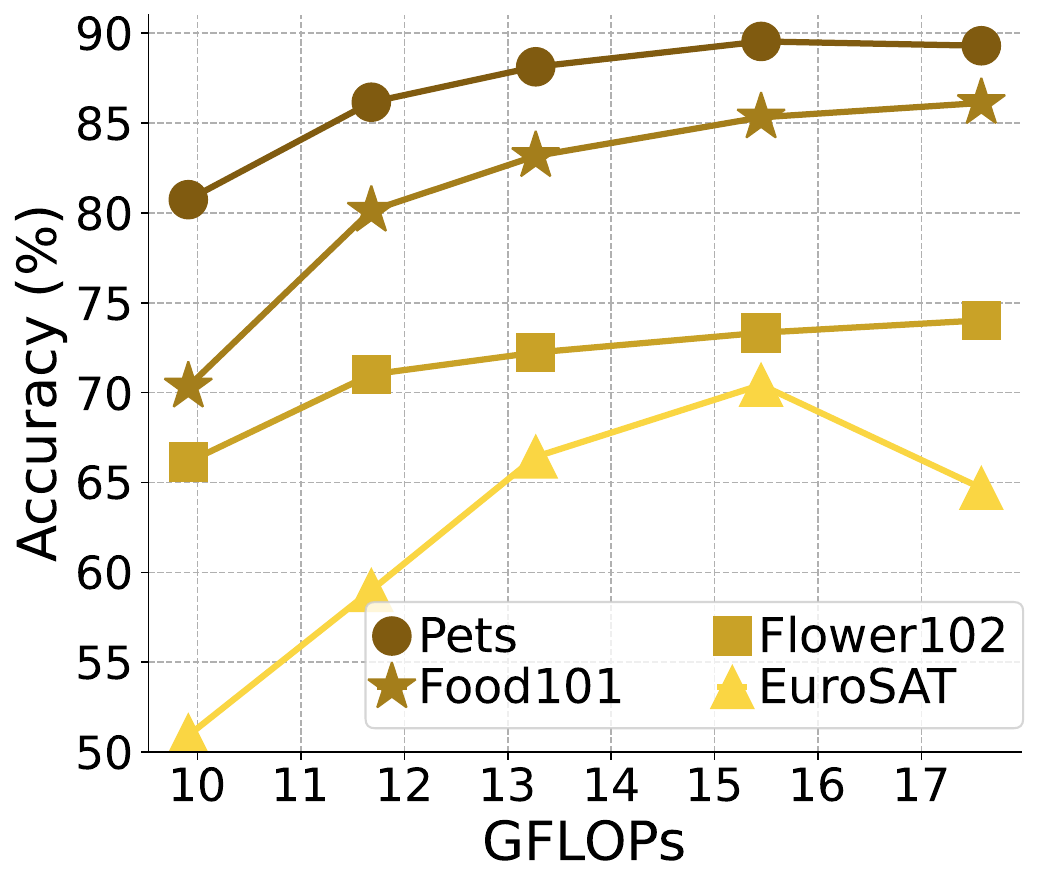}\vspace{-1ex}
        \caption{Impact of GFLOPs budget.}
        \label{abl.rate}
    \end{subfigure}
    \vspace{-1.5ex}
    \caption{Impact of reservoir size and GFLOPs budgets.}\vspace{-2ex}
    \label{fig:abl.size.budget.tda}
\end{figure}
\subsection{Ablation Study}\label{sec:ablation}
We conducted a comprehensive ablation study to evaluate TCA's effectiveness and efficiency. For further analysis on hyperparameter, see the supplementary material.

\noindent \textbf{Results on Various Severity Levels. }To evaluate the effectiveness of TCA across different shift levels, we conduct experiments on five severity levels of three corruption types in the CIFAR-100-Corrupted dataset. Our results in \cref{tab.cifar} reveal a performance decline for EViT across most severity levels, particularly for snow and brightness corruptions. This may be due to the CLIP model embedding perturbed samples differently when token reduction is applied, leading to misinterpretations. In contrast, our proposed TCA consistently demonstrates improvements, highlighting the effectiveness of the DTR module in preserving informative and reliable token representations for logits correction.

\begin{table}[h]
  \centering
  \Large
  \caption{Improvements over CLIP inference on CIFAR-100-C.}\vspace{-1.5ex}
  \label{tab.cifar}
  \resizebox{1\linewidth}{!}{
    \begin{tabular}{c ccc ccc ccc}  
      \toprule
      & \multicolumn{3}{c}{$\operatorname{\textbf{Contrast}}$} 
      & \multicolumn{3}{c}{$\operatorname{\textbf{Snow}}$} 
      & \multicolumn{3}{c}{$\operatorname{\textbf{Brightness}}$} \\ 
      \cmidrule(lr){2-4} \cmidrule(lr){5-7} \cmidrule(lr){8-10}  
      $\operatorname{Severity}$ & $\operatorname{CLIP}$ & $\operatorname{EViT}$ & $\operatorname{Ours}$ 
      & $\operatorname{CLIP}$ & $\operatorname{EViT}$ & $\operatorname{Ours}$ 
      & $\operatorname{CLIP}$ & $\operatorname{EViT}$ & $\operatorname{Ours}$ \\
      \midrule
      \midrule
      \textbf{1} & 31.90 & {\cellcolor{blue!10}-3.17\%} & \textbf{\cellcolor{pink!30}+2.16\%} 
      & 35.34 & {\cellcolor{blue!10}-2.49\%} & \textbf{\cellcolor{pink!30}+1.50\%} 
      & 41.00 & {\cellcolor{blue!10}-2.10\%} & \textbf{\cellcolor{pink!30}+2.10\%} \\
      \textbf{2} & 20.67 & {\cellcolor{pink!30}+0.68\%} & \textbf{\cellcolor{pink!30}+3.58\%} 
      & 29.72 & {\cellcolor{blue!10}-1.99\%} & \textbf{\cellcolor{pink!30}+1.14\%} 
      & 41.44 & {\cellcolor{blue!10}-1.54\%} & \textbf{\cellcolor{pink!30}+2.85\%}\\
      \textbf{3} & 15.05 & {\cellcolor{blue!10}-1.53\%} & \textbf{\cellcolor{pink!30}+7.04\%} 
      & 29.14 & {\cellcolor{blue!10}-1.68\%} & \textbf{\cellcolor{pink!30}+1.61\%} 
      & 41.83 & {\cellcolor{blue!10}-1.82\%} & \textbf{\cellcolor{pink!30}+2.37\%} \\
      \textbf{4} & 8.85 & {\cellcolor{blue!10}-0.34\%} & \textbf{\cellcolor{pink!30}+11.41\%} 
      & 27.04 & {\cellcolor{blue!10}-2.51\%} & \textbf{\cellcolor{pink!30}+1.48\%} 
      & 41.12 & {\cellcolor{blue!10}-1.82\%} & \textbf{\cellcolor{pink!30}+4.74\%} \\
      \textbf{5} & 2.69 & {\cellcolor{pink!30}+1.49\%} & \textbf{\cellcolor{pink!30}+18.59\%} 
      & 24.85 & {\cellcolor{blue!10}-0.80\%} & \textbf{\cellcolor{pink!30}+3.30\%} 
      & 38.10 & {\cellcolor{blue!10}-2.13\%} & {\cellcolor{pink!30}\textbf{+4.75\% }}\\
      \bottomrule
    \end{tabular}}\vspace{-2ex}
\end{table}

\noindent\textbf{Results on Multimodal VLMs. }To evaluate the generalizability of our proposed method, we conduct additional experiments on SigLIP and SigLIP v2. As shown in \cref{tab:siglip}, TCA consistently improves the performance of both SigLIP and SigLIP v2 without requiring model tuning. Notably, EuroSAT achieves a significant gain of 21.12\% over SigLIP direct inference, aligning with trends observed in \cref{tab:cd-exp}. On average, TCA yields improvements of 2.21\% for SigLIP and 1.32\% for SigLIP v2 across datasets, demonstrating its potential. While minor degradations appear in Food101, overall performance remains stable or improved across most datasets, reinforcing TCA's robustness.

\begin{table*}[t]
\caption{Improvements in Cross-Dataset benchmark over SigLIP and SigLIP v2 inference on ViT-B/16.}\vspace{-1.5ex}
  \centering
  \resizebox{\linewidth}{!}{
    \begin{tabular}{l*{11}{c}}
      \toprule
      \textbf{Method} & {Aircraft} & {Caltech101} & {Cars} & {DTD} & {EuroSAT} & {Flower102} & {Food101} & {Pets} & {SUN397} & {UCF101} & {\textbf{Average}} \\
      \midrule
      \midrule
      SigLIP & 40.50 & 97.44 & 90.71 & 62.83 & 39.86 & 84.13 & 89.06 & 93.10 & 69.65 & 70.84 & 73.81 \\
      \textbf{+TCA$_{R=0.9}$} & \cellcolor{pink!30}+0.74\% & \cellcolor{pink!30}+0.58\% & \cellcolor{pink!30}+0.12\% & \cellcolor{pink!30}+2.44\% & \cellcolor{pink!30}+21.12\% & \cellcolor{pink!30}+1.39\% & \cellcolor{blue!10}-0.17\% & \cellcolor{pink!30}+0.56\% & \cellcolor{pink!30}+2.02\% & \cellcolor{pink!30}+3.40\% & \cellcolor{pink!30}+2.21\%\\
      \midrule
      SigLIP v2  & 50.05 & 97.61 & 93.36 & 62.41 & 42.09 & 85.30 & 90.14 & 94.88 & 73.34 & 72.09 & 76.13 \\
      \textbf{+TCA$_{R=0.9}$} & \cellcolor{pink!30}+0.42\% & \cellcolor{pink!30}+0.25\% & \cellcolor{pink!30}+0.11\%  & \cellcolor{pink!30}+2.76\% & \cellcolor{pink!30}+11.17\% & \cellcolor{pink!30}+0.39\% & \cellcolor{blue!10}-0.04\% & \cellcolor{pink!30}+0.02\% & \cellcolor{pink!30}+0.64\% & \cellcolor{pink!30}+3.22\% & \cellcolor{pink!30}+1.32\% \\
      \bottomrule
    \end{tabular}}\vspace{-2ex}
  \label{tab:siglip}
\end{table*}

\noindent\textbf{Impact of GFLOPs Budget.} We evaluate TCA under different GFLOPs budgets: $R=\{0.6, 0.7, 0.8, 0.9\}$, resulting in GFLOPs of 9.91, 11.68, 13.27, and 15.45, respectively, compared to the baseline ($R=1$, 17.58 GFLOPs). As shown in \cref{abl.rate}, condensing inattentive tokens can even enhance performance on certain datasets, notably Pets, and EuroSAT. Specifically for EuroSAT, when $R=0.9$, the model's adaptation performance is significantly improved, aligning with our findings in \cref{fig:leave-one-out}. However, excessively aggressive pruning budgets (\textit{e.g.,} GFLOPs less than 13) lead to significant performance degradation across all datasets. This occurs since higher pruning rates may inadvertently remove informative tokens, causing irreversible harm in training-free scenarios where we lack the ability to update the model for further correction.

\begin{table}
\centering
\caption{Impact of reservoir updating strategy with CLIP.}
\vspace{-1.5ex}
\resizebox{0.96\linewidth}{!}{%
\begin{tabular}{l  c c c c}
\toprule
   & \textbf{FIFO} &\textbf{Uncertainty} & \textbf{Similarity-enf} & \textbf{Diversity-enf}\\ 
\midrule
\midrule
Pets & 89.21 & 89.18  & 88.91 & \cellcolor{pink!30}\textbf{89.53} \\
Flower102 & 70.65 & 73.28  & 72.15 & \cellcolor{pink!30}\textbf{73.33} \\
EuroSAT & 50.28 & 70.20  & 68.48 & \cellcolor{pink!30}\textbf{70.43} \\
\bottomrule
\end{tabular}}\vspace{-1ex}
\label{tab.cache_saving_strategy}
\end{table}

\noindent\textbf{Impact of Reservoir Saving Strategy.} In \cref{tab.cache_saving_strategy}, we examine the performance changes across different reservoir saving strategies. We compare several approaches: \textit{First-In-First-Out (FIFO)}; an \textit{uncertainty-based} strategy, which discards the most uncertain sample when the reservoir reaches capacity; a \textit{similarity-enforced} strategy, where samples with high certainty and high cosine similarity to the saved samples are preferred; and a \textit{diversity-enforced} strategy, which prioritizes saving prototypes that contain distinct tokens compared to those already stored. Our results show that the FIFO strategy performs poorly on Flower102 and EuroSAT, likely because CLIP's low confidence leads to retaining misclassified samples. Conversely, Pets has high CLIP zero-shot accuracy (86.91\% in \cref{tab:cd-exp}), which makes FIFO acceptable. Among all strategies, the diversity-based approach consistently achieves the best performance. This is intuitive, as it maintains a representative set of features by capturing dataset diversity, whereas entropy-based methods may store homogenous features and overlook multiple class prototypes. By prioritizing diversity, our method ensures that a more representative set of features is maintained, leading to more robust performance across datasets. 

\noindent\textbf{Impact of Reservoir Size $M$.}
We assess the effectiveness of TCA across various reservoir sizes \( M \) on Pets, Flower102, and EuroSAT datasets, as illustrated in \cref{abl.cache_size}. Remarkably, although the best performances are achieved at different reservoir sizes for different datasets, our TCA consistently maintains stable and high performance across a wide range of \( M \) values. This showcases the robustness and flexibility of TCA with respect to different reservoir budgets. Notably, even under extreme conditions with a minimal reservoir size (\textit{i.e.,} \( M=1 \)), our strategy significantly surpasses the strongest baseline method, TDA, by a large proportion on the EuroSAT dataset (14.2\%). 

\begin{table}
\centering
\caption{Ablation study of the proposed components in TCA.}
\vspace{-1.5ex}
\resizebox{0.67\linewidth}{!}{%
\begin{tabular}{c c c c c }
\toprule
\textbf{$\mathbf{A}_{c^*}^l$} & \textbf{$\mathbf{S}^\text{head}$} & \textbf{Food101} & \textbf{Pets} & \textbf{EuroSAT}\\
 \midrule
 \midrule
  & & 85.25 & 88.77 & 68.14\\
  \textcolor{tickgreen}{\ding{51}} &  & 85.31 & 88.96 & 67.69 \\
  & \textcolor{tickgreen}{\ding{51}} & 85.25 & 89.23 & 67.14 \\
  \midrule
  \textcolor{tickgreen}{\ding{51}} & \textcolor{tickgreen}{\ding{51}} & \cellcolor{pink!30}\textbf{85.31} & \cellcolor{pink!30}\textbf{89.53} & \cellcolor{pink!30}\textbf{70.43} \\
  \bottomrule
\end{tabular}}
\label{table:ablation}\vspace{-2ex}
\end{table}

\noindent\textbf{Impact of Component.} The impact of the domain anchor tokens $\mathbf{A}_{c^*}^l$ and the head-wise sorting score $\mathbf{S}^\text{head}$ (\cref{sec:pruning}) is presented in \cref{table:ablation}. We observe that each component individually contributes to performance improvements. On the Food101 and Pets datasets, incorporating either component yields measurable gains in accuracy. By leveraging historical domain anchor tokens, the model acquires rich contextual information, enhancing the stability of token importance over time. Simultaneously, cross-head token sorting ensures that token pruning decisions are more robust by accounting for consensus across attention heads. An intriguing case arises with the EuroSAT dataset. Here, the baseline performance without any components is 68.14\%. Applying either component alone results in a slight performance decrease. However, when both components are used together, performance significantly improves to 70.43\%. This outcome emphasizes the necessity of combining historical domain anchor tokens and cross-head token sorting to fully realize the model's potential.

\begin{figure}
        \centering
        \includegraphics[width=0.85\linewidth]{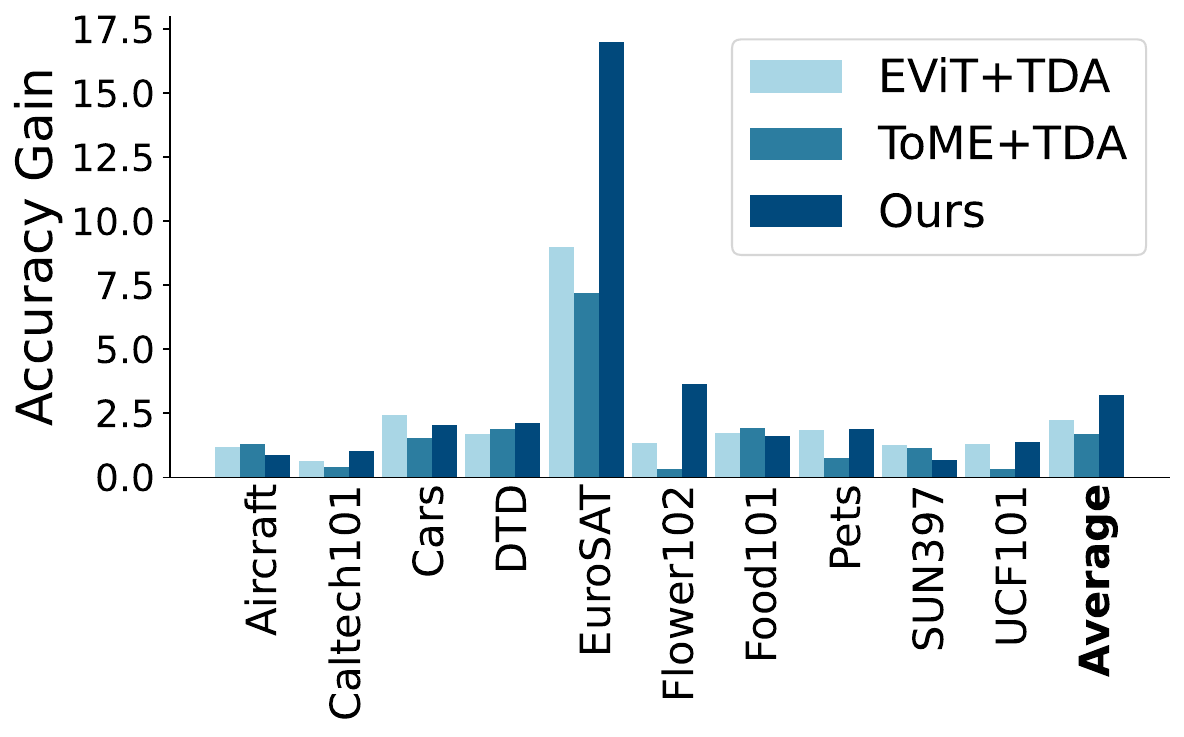}\vspace{-1.5ex}
        \caption{Accuracy gain of TCA and TDA + EViT/ToME over TDA combined with ATS.}
        \label{abl.tda_prune}\vspace{-2ex}
    \end{figure} 
\noindent \textbf{Comparison with TDA $+$ Token Condensation. }We evaluate the performance for TDA $+$ token pruning and merging baselines and show the performance gain over TDA $+$ ATS in \cref{abl.tda_prune}. Although TDA achieves considerable performance gain, it heavily relies on the negative cache and a large set of hyperparameters. In contrast, TCA's accuracy gain significantly surpasses that of TDA $+$ EViT and TDA $+$ ToME across multiple datasets and on average, even with a minimal set of hyperparameters, highlighting its superior adaptation capability.

\begin{figure}
    \centering 
    \includegraphics[width=0.95\linewidth]{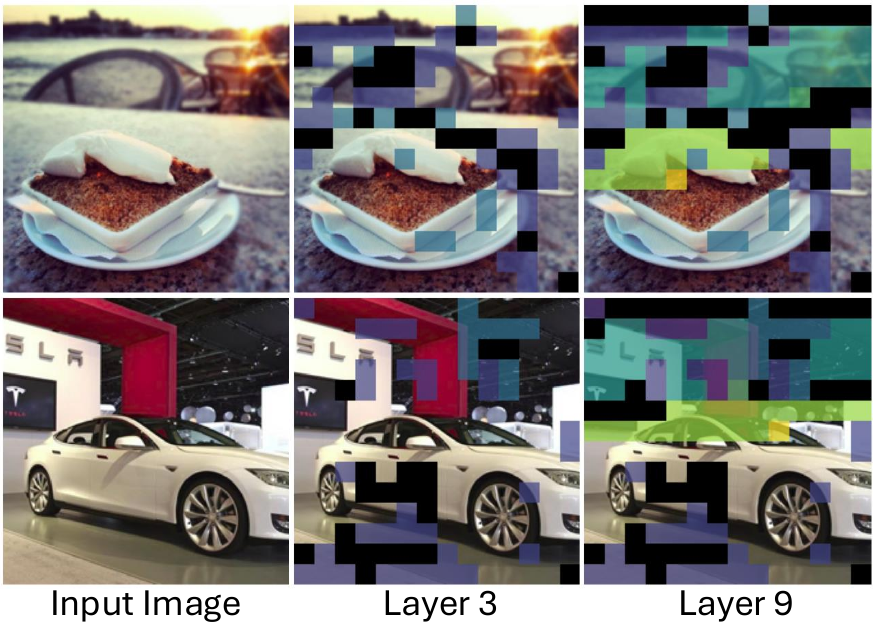}\vspace{-2ex}
    \caption{Examples of our token condensation. More visualizations can be found in the supplementary material.}
    \label{fig:token_viz_main}\vspace{-2ex}
\end{figure}
\noindent\textbf{Visualization of Token Condensation.} We visualize the pruned and merged tokens of different ViT layers in \cref{fig:token_viz_main}. The black mask indicates pruned regions while different colors are set for different merging clusters. As token condensation progresses, non-discriminative tokens are gradually removed, leading to better alignment with the text semantics. See the supplementary material for more details.

%% file: sec/5_conclusion.tex
\vspace{-3ex}
\section{Conclusion}
\vspace{-1ex}
We introduced Token Condensation as Adaptation (TCA), a novel training-free test-time adaptation method on VLMs such as CLIP. Our experiments across various VLMs demonstrated that token condensation significantly enhances visual-text alignment while also serving as an interpretation of visual semantics. Besides, TCA reduces GFLOPs as a beneficial byproduct, improving computational efficiency. Despite our empirical findings, a rigorous theory of CLIP's generalizability remains underexplored. To bridge the gap, we provide a theoretical analysis, examine TCA's applicability to other VLMs, and acknowledge its limitations in the supplementary material.

%% file: sec/x_supp.tex
\clearpage
\setcounter{page}{1}
\maketitlesupplementary

\noindent This supplementary material provides additional details of TCA, including method descriptions, theoretical analysis, empirical results, and the algorithm. We also discuss TCA's applicability and limitations. To further illustrate the method, we include visual aids for token condensation.

\begin{itemize}
    \item \textbf{\cref{sec.coreset}:} Details of the Coreset Selection Strategy.
    \item \textbf{\cref{sec. theory}:} Theoretical Analysis.
    \item \textbf{\cref{sec:add_exp}:} Additional Experiments and Ablation Study.
    \item \textbf{\cref{sec:alg}:} Token Condensation Algorithm.
    \item \textbf{\cref{sec:q_stu}:} Quantitative Study ($R=0.7$).
    \item \textbf{\cref{sec.general}:} Discussion on TCA's Generalizability.
    \item \textbf{\cref{sec:limitation}:} Potential Limitation of TCA.
\end{itemize}
\subsection{Details of Coreset Selection}\label{sec.coreset}
In domain-aware token merging, we first identify the most representative tokens $\hat{\mathbf{V}}^l_{\operatorname{merge}}\in\mathbb{R}^{K\times D_v}$ from $\mathbf{V}^l_{\Phi}$ and assigns the remaining ambiguous tokens to these selected tokens. This strategy is equivalent to solving the K-Center problem \citep{DBLP:journals/gis/Wolf11, DBLP:conf/iclr/SenerS18}. The objective is to select $K$ center tokens such that the maximum distance between any token and its nearest center is minimized. The greedy search for coreset optimization is defined as follows:
\begin{equation}
\mathbf{C}^* = {\arg\min}_{\mathbf{C} \subseteq \mathbf{V}^l_{\Phi}, |\mathbf{C}| = K} \max_{\mathbf{v}_i^l \in \mathbf{V}^l_{\Phi}} \min_{\mathbf{v}_c^l \in \mathbf{C}} d(\mathbf{v}_i^l, \mathbf{v}_c^l),
\end{equation}
where $\mathbf{C}^*\in\mathbb{R}^{K\times D_v}$ represents the set of selected center tokens, $K$ is the number of centers, and $d(\cdot, \cdot)$ is the distance metric between token $\mathbf{v}_i^l$ and center token $\mathbf{v}_c^l$. Once the center tokens $\mathbf{C}^*$ are selected, the remaining tokens are assigned to their nearest centers, and the ambiguous tokens are merged as: \begin{equation}\label{eq:token_merged}
\hat{\mathbf{V}}_{\operatorname{merged}}^l = \frac{1}{|\mathcal{N}(k)|} \sum\nolimits_{\mathbf{v}_i^l \in \mathcal{N}(k)} \mathbf{v}_i^l,
\end{equation}
where $\mathcal{N}(k)$ represents the set of tokens assigned to center $k$. The value of $K$ is kept small, with $K \ll N$, allowing our merging algorithm to operate with linear complexity.

\subsection{Theoretical Analysis}\label{sec. theory}
The theoretical foundations of CLIP's generalization remain underexplored, with ongoing debates on whether it arises from train-test similarity \cite{DBLP:conf/iclr/MayilvahananWRB24}, spurious feature reliance \cite{DBLP:conf/nips/WangL0S0024}, or other factors. While we did not include rigorous proof, we connect our TCA to \textit{PAC-Bayesian generalization theory}. We model token selection as a stochastic hypothesis, where the \textit{posterior} $\mathbb{Q}$ over retained tokens follows a \textit{Gibbs formulation}, favoring subsets that minimize cosine similarity variance with texts:
\begin{equation}\footnotesize
\begin{split}\nonumber 
     \mathbb{Q}(\hat{\mathbf{V}}) &= \frac{1}{Z}\exp\left(-\lambda \mathrm{Var}\left(\operatorname{cos}(\mathbf{V}, \mathbf{t}_c)\right)\right),\\ 
     \mathbb{E}_{\operatorname{ood}}[-\operatorname{cos}(\hat{\mathbf{V}}, \mathbf{t}_c)]\leq& \mathbb{E}_{\operatorname{id}}[-\operatorname{cos}(\mathbf{V}, \mathbf{t}_c)] + \sqrt{\frac{1}{2}(D_{\operatorname{KL}}(\mathbb{Q}\|\mathbb{P})+\log\frac{m}{\delta}}.
\end{split}
\end{equation}
This supports the PAC-Bayes bound, where TCA improves generalization by reducing KL divergence between test-time token selection and CLIP's inaccessible pretraining distribution, which we approximate using DTR. Empirical results in \cref{fig:anchor2text} confirm this, showing that retained tokens act as stable anchors for text alignment.

\subsection{Additional Results}\label{sec:add_exp}
\noindent\textbf{Impact of Visual Backbone. }
\begin{table*}[ht]
  \centering
  \caption{Results on the cross-dataset benchmark with CLIP ViT-L/14. $^*$ denotes the averaged GFLOPs across all datasets.}
  \resizebox{\linewidth}{!}{
    \begin{tabular}{l*{11}cl}
      \toprule
      \textbf{Method} & Aircraft & Caltech101 & Cars & DTD & EuroSAT & Flower102 & Food101 & Pets & SUN397 & UCF101 & \textbf{Average} & \textbf{GFLOPs}\\
      \midrule
      \midrule
      CLIP & 31.59 & 94.56 & 78.12 & 57.03 & 63.00 & 79.58 & 90.92 & 93.46 & 69.05 & 76.13 & \cellcolor{pink!30}{73.34} & \cellcolor{blue!10}{81.14}\\
       Tent & 27.45 & 94.97 & 76.93 & 57.15 & 66.20 & 74.83 & 89.20 & 93.27 & 68.73 & 75.73 & \cellcolor{pink!30}{72.45} & \cellcolor{blue!10}{81.14} \\
      SAR & 26.07 & 94.52 & 75.58 & 56.91 & 63.77 & 75.03 & 89.13 & 93.05 & 68.39 & 75.50 & \cellcolor{pink!30}{71.80} & \cellcolor{blue!10}{81.14} \\
      TPT & 30.06 & 95.21 & 76.84 & 52.30 & 55.11 & 76.21 & 88.56 & 93.08 & 67.69 & 73.78 & \cellcolor{pink!30}{70.88} & \cellcolor{blue!10}{143.31}\\
      TDA & 33.42 & 95.46 & 78.72 & 57.39 & 66.27 & 79.94 & 90.83 & 93.27 & 70.74 & 78.14 & \cellcolor{pink!30}{74.42} & \cellcolor{blue!10}{81.14}\\
      \midrule
      EViT$_{R=0.9}$ & 31.23 & 94.56 & 76.59 & 56.38 & 63.04 & 79.13 & 90.08 & 93.32 & 68.54 & 76.40 & \cellcolor{pink!30}{72.93} & \cellcolor{blue!10}{65.19} \\
      ToME$_{R=0.9}$ & 28.29 & 92.54 & 71.26 & 56.68 & 60.30 & 77.87 & 89.77 & 91.28 & 68.21 & 72.22 & \cellcolor{pink!30}{70.84} & \cellcolor{blue!10}{64.74} \\
      ATS$_{R=0.9}$ & 25.74 & 93.39 & 67.69 & 55.02 & 52.81 & 76.78 & 86.48 & 91.50 & 66.26 & 72.56 & \cellcolor{pink!30}{68.82} & \cellcolor{blue!10}{43.62$^*$}\\
      EViT$_{R=0.7}$ & 26.94 & 92.94 & 62.55 & 53.96 & 52.04 & 73.24 & 80.69 & 90.00 & 63.70 & 71.21 & \cellcolor{pink!30}{66.73} & \cellcolor{blue!10}{40.78} \\
      ToME$_{R=0.7}$ & 15.60 & 83.73 & 38.43 & 49.82 & 44.51 & 59.36 & 72.65 & 77.73 & 58.32 & 50.99 & \cellcolor{pink!30}{55.11} & \cellcolor{blue!10}{40.05} \\
      ATS$_{R=0.7}$ & 6.87 & 67.87 & 16.37 & 40.78 & 30.12 & 37.43 & 34.50 & 60.94 & 30.07 & 33.44 & \cellcolor{pink!30}{35.84} & \cellcolor{blue!10}{26.76$^*$}\\
      \midrule
      \textbf{TCA}$_{R=0.9}$ & 33.84 & 96.39 & 76.93 & 56.38 & 67.74 & 80.71 & 90.21 & 93.54 & 70.02 & 78.24 & \cellcolor{pink!30}{74.40} & \cellcolor{blue!10}{\(\mathbf{65.24}_{\mathbf{\textcolor{blue}{-19.6\%}}}\)}\\
      \textbf{TCA}$_{R=0.7}$ & 29.73 & 94.81 & 63.72 & 53.72 & 60.69 & 76.00 & 81.55 & 90.02 & 65.61 & 73.14 & \cellcolor{pink!30}{68.90} & \cellcolor{blue!10}{\(\mathbf{41.44}_{\mathbf{\textcolor{blue}{-48.9\%}}}\)}\\
      \bottomrule
    \end{tabular}
  } 
  \label{tab:cd-exp-vitl}
\end{table*}
Trends similar to ViT-B/16 are observed with the \textbf{ViT-L/14} architecture, as shown in \cref{tab:cd-exp-vitl}. TCA consistently surpasses TDA across multiple datasets, including Aircraft, Caltech101, EuroSAT, Flower102, Pets, and UCF101, while adhering to a limited GFLOPs budget (19.6\% GFLOPs reduction). Even with a 48.9\% reduction in GFLOPs, TCA continues delivering satisfactory results. This demonstrates the scalability and robustness of our method across different model sizes, reinforcing its effectiveness without additional training.

\begin{table}[H]
\centering
\caption{Impact of scale factor $\beta$.}
\resizebox{0.8\linewidth}{!}{%
\begin{tabular}{l  c c c c c}
\toprule
\textbf{$\beta$} &\textbf{0.01} & \textbf{0.05} & \textbf{1} & \textbf{3} & \textbf{5} \\ 
\midrule
\midrule
Pets & 89.51 & \textbf{89.53} & 89.37 & 89.42 & 89.26 \\
Flower102 & \textbf{73.33} & 73.08 & 70.93 & 70.56 & 70.44 \\
EuroSAT & 63.64 & 64.06 & 69.86 & 70.26 & \textbf{70.43} \\
\bottomrule
\end{tabular}}
\label{tab.scale}
\end{table}

\noindent\textbf{Impact of Logits Correction Temperature $\beta$. }
 \cref{tab.scale} examines how different logits correction temperatures $\beta$ affect the adaptation results. The intuition is that with a smaller $\beta$ value, the logits correction will emphasize the tokens in shallower layers (\cref{eq:logit_correction}), while a larger $\beta$ value will shift the focus to deeper layers. We observe that a smaller value of $\beta$ is preferred for the Pets dataset as it contains animals as objects, requiring more high-level contextual information for accurate predictions \cite{DBLP:conf/nips/RaghuUKZD21}. In contrast, for EuroSAT, the best predictions are obtained with larger $\beta$ values, suggesting that low-level, local information is crucial. This aligns well with the nature of the dataset, where different types of land can be distinguished by features such as colors and edges.   Nevertheless, our method consistently provides significant improvements across all $\beta$ values, with accuracy gains of up to 20\%, highlighting the effectiveness of logits correction using the domain anchor tokens.
 
 \begin{table}[H]
\centering
\noindent\caption{Impact of correction weight $\lambda$.}
\resizebox{\linewidth}{!}{%
\begin{tabular}{l  c c c c c c c}
\toprule
\textbf{$\lambda$} & \textbf{2} & \textbf{3} & \textbf{4} & \textbf{5} & \textbf{6} & \textbf{7} & \textbf{8} \\ 
\midrule
\midrule
Pets & \textbf{89.53} & 89.32 & 89.13 & 88.96 & 88.96 & 88.66 & 88.44\\
Flower102 & 72.43 & 72.76 & 73.20 & 73.16 & 73.16 & \textbf{73.33} & 73.16 \\
EuroSAT & 60.15 & 65.74 & 68.80 & 69.51 & 69.84 & 70.16 & \textbf{70.43} \\
\bottomrule
\end{tabular}}
\label{tab.correction_weight}
\end{table} 

\noindent\textbf{Impact of Correction Weight $\lambda$. }To investigate how different correction weights $\lambda$ affect performance, as described in \cref{eq:logit_correction}, we conducted experiments across a wide range of $\lambda$ values, from 2 to 8, as shown in \cref{tab.correction_weight}. We observe that Pets exhibits stable results across different $\lambda$ values, indicating that less aggressive correction is sufficient.
In contrast, datasets such as Flower102 and EuroSAT which initially do not perform well on CLIP, benefit from stronger corrections, achieving their best performance with larger correction weights of 7 and 8, respectively. This highlights the effectiveness of our logits correction module.

\begin{table}[H]
\centering
\caption{Impact of token merging/pruning ratio.}
\resizebox{0.68\linewidth}{!}{%
\begin{tabular}{l  c c c}
\toprule
\textbf{Merging:Pruning} & \textbf{0:1} & \textbf{1:2} & \textbf{2:1} \\ 
\midrule
\midrule
Pets & 89.04 & 88.99 & \cellcolor{pink!30}\textbf{89.53} \\
EuroSAT & 69.63 & 69.98 & \cellcolor{pink!30}\textbf{70.43} \\
\bottomrule
\end{tabular}}
\label{tab.merge:prune}
\end{table}
\noindent \textbf{Impact of Pruning \& Merging Ratio. }
We experiment with different token pruning and merging ratios under the same computational budget, as shown in \cref{tab.merge:prune}. Incorporating token diversity through merging consistently enhances performance. Specifically, the 2:1 merging-to-pruning ratio outperforms other configurations, especially those favoring pruning. This is because merging preserves diverse token representations by K coresets that pure pruning might discard. When comparing pruning-only (0:1) with the 1:2 merging-pruning ratio on Pets, pruning-only performs better. This may be because the dataset features images with a single prominent object, meaning that pruning background tokens has minimal impact since essential object information remains intact. In contrast, for the EuroSAT dataset, which comprises diverse satellite imagery, simply pruning tokens leads to the loss of important contextual features necessary for accurate classification.

\begin{table}[h]
\centering
\caption{Impact of the merging center number K.}
\resizebox{0.7\linewidth}{!}{%
\begin{tabular}{l c c c c}
\toprule
\textbf{K} & \textbf{1} & \textbf{2} & \textbf{3} & \textbf{4} \\ 
\midrule
\midrule
Pets & 89.29 & \cellcolor{pink!30}{\textbf{89.53}} & 89.29 & 89.21 \\
EuroSAT & 66.25 & \cellcolor{pink!30}{\textbf{70.43}} & 66.96 & 67.44\\
Food101 & 85.15 & \cellcolor{pink!30}{\textbf{85.31}} & 85.31 & 85.38\\
\bottomrule
\end{tabular}}
\label{tab.center_k}
\end{table}
\noindent \textbf{Impact of Merging Center Number $K$. }We evaluate TCA performance by giving different numbers of merging centers \( K \) for Pets, EuroSAT, and Food101 datasets. As shown in \cref{tab.center_k}, setting $K = 2$ consistently yields the best results. This choice balances preserving important information and reducing redundancy. A smaller $K$ (\textit{i.e.,} $K = 1$) may oversimplify the merging process, leading to the loss of critical details, especially in diverse datasets like EuroSAT. Conversely, increasing $K$ beyond 2 introduces unnecessary complexity and can over-segment the token space, retaining redundant tokens that contribute little to classification. Therefore, maintaining a very small $K$ (where $K \ll N$) is sufficient and advantageous.

\begin{table*}[t]
  \centering\caption{Results on the out-of-distribution benchmark with CLIP ViT-B/16. $^*$ denotes the averaged GFLOPs across all datasets.}
  \resizebox{\linewidth}{!}{
    \begin{tabular}{l*{9}c}
      \toprule
      Method & Aug-free &ImageNet & ImageNet-A & ImageNet-V2 & ImageNet-R & ImageNet-S & \textbf{Average} & \textbf{OOD Average} & \textbf{GFLOPs}\\
      \midrule
      \midrule
      CLIP & $\checkmark$ & 68.34 & 49.89 & 61.88 & 77.65 & 48.24 & \cellcolor{pink!30}{61.20} & \cellcolor{pink!30}{59.42} & \cellcolor{blue!10}{17.59}\\
      \midrule
      Tent & \textcolor{tickgreen}{\ding{51}} & 65.49 & 44.57 & 59.26 & 78.72 & 22.52 & \cellcolor{pink!30}{54.11} & \cellcolor{pink!30}{51.27} & \cellcolor{blue!10}{17.59}\\
      SAR & \textcolor{tickgreen}{\ding{51}} & 58.52 & 33.71 & 53.95 & 76.08 & 39.24 & \cellcolor{pink!30}{52.30} & \cellcolor{pink!30}{50.75} & \cellcolor{blue!10}{17.59}\\
      TPT & \textcolor{crossred}{\ding{55}} & 68.98 & 54.77 & 63.45 & 77.06 & 47.94 & \cellcolor{pink!30}{62.44} & \cellcolor{pink!30}{60.81} & \cellcolor{blue!10}{1108.61}\\
      Diff-TPT & \textcolor{crossred}{\ding{55}} & 70.30 & 55.68 & 65.10 & 75.00 & 46.80 & \cellcolor{pink!30}{62.28} & \cellcolor{pink!30}{60.52} & \cellcolor{blue!10}{-} \\
      C-TPT & \textcolor{crossred}{\ding{55}} & 69.30 & 52.90 & 63.40 & 78.00 & 48.50 & \cellcolor{pink!30}{62.42} & \cellcolor{pink!30}{60.70} & \cellcolor{blue!10}{1108.61}\\
      MTA & \textcolor{crossred}{\ding{55}} & 70.08 & 58.06 & 64.24 & 78.33 & 49.61 & \cellcolor{pink!30}{64.06} & \cellcolor{pink!30}{62.56} & \cellcolor{blue!10}{-} \\
      TDA & \textcolor{tickgreen}{\ding{51}} & 69.26 & 50.82 & 62.23 & 77.93 & 50.26 & \cellcolor{pink!30}{62.10} & \cellcolor{pink!30}{60.31} & \cellcolor{blue!10}{17.59} \\
      \midrule
      EViT$_{R=0.95}$ & \textcolor{tickgreen}{\ding{51}} & 68.32 & 49.46 & 61.73 & 77.00 & 47.76 & \cellcolor{pink!30}{60.85} & \cellcolor{pink!30}{58.99} & \cellcolor{blue!10}{16.31} \\
      ToME$_{R=0.95}$ & \textcolor{tickgreen}{\ding{51}} & 67.57 & 48.81 & 60.88 & 75.78 & 47.05 & \cellcolor{pink!30}{60.02} & \cellcolor{pink!30}{58.13} & \cellcolor{blue!10}{16.21} \\
      ATS$_{R=0.95}$ & \textcolor{tickgreen}{\ding{51}} & 65.83 & 49.80 & 59.47 & 71.09 & 43.38 & \cellcolor{pink!30}{57.91} & \cellcolor{pink!30}{55.94} & \cellcolor{blue!10}{11.50$^*$} \\
      \midrule
      \textbf{TCA}$_{R=0.95}$ & \textcolor{tickgreen}{\ding{51}} & 68.88 & 50.13 & 62.10 & 77.11 & 48.95 & \cellcolor{pink!30}{61.43} & \cellcolor{pink!30}{59.57} & \cellcolor{blue!10}{16.55} \\
      \bottomrule
    \end{tabular}
  } 
  \label{tab:ood-exp-vit-B}
\end{table*}

\noindent \textbf{Impact of Benchmark Datasets. }
We conducted experiments on the OOD benchmark which focuses on evaluating the model's effectiveness on shifted data using label sets previously seen by CLIP. This includes variants of ImageNet \cite{DBLP:conf/cvpr/DengDSLL009}: ImageNet-A \cite{DBLP:journals/corr/abs-1907-07174}, ImageNet-V2 \cite{DBLP:conf/icml/RechtRSS19}, ImageNet-R \cite{DBLP:conf/iccv/HendrycksBMKWDD21}, and ImageNet-S \cite{DBLP:conf/nips/WangGLX19}. A consistent observation can be seen in the out-of-distribution (OOD) benchmark, where TCA demonstrates significant improvements over the CLIP baseline under a constrained GFLOPs budget of $R=0.95$, as shown in \cref{tab:ood-exp-vit-B}. TCA outperforms traditional test-time adaptation methods while maintaining efficiency. TCA also achieves superior results on ImageNet-R and ImageNet-S, outperforming TPT without augmentation. Additionally, when compared to other training-based approaches, even those with unlimited computational budgets, TCA delivers comparable performance. However, we observe that TCA does not perform as strongly on the OOD benchmark as it does on the CD benchmark even with a higher rate $R$. This may be due to the conceptual shifts in OOD datasets, as shown in \cref{sec:limitation}, which could present a challenge for training-free adaptation methods.

\subsection{Algorithm} \label{sec:alg}
\begin{algorithm}[t]
\caption{Token Condensation at the $l$-Layer in $E_v$}\label{alg:ours}
\begin{algorithmic}[1]
\Require \\
    Token reservoir $\mathfrak{R}$; \\
    Visual patches $\mathbf{V}^{l-1}$ at layer $l-1$; \\
    Pruning threshold $\theta_{\text{prune}}(\alpha \cdot R)$; \\
    Merging threshold $\theta_{\text{merge}}(R)$
\Ensure
    Token-efficient visual feature $\hat{\mathbf{V}}^{l}$
\State \textbf{Domain Anchor Token Selection}: Obtain $\mathbf{A}^{l-1}_{c^*}$, using domain anchor tokens in $\mathbf{\mathfrak{R}}$ and sample's \texttt{<cls>} token $\mathbf{v}^{l}_{\text{cls}}$
\State Compute cross-head scores $\mathbf{S}_i^\text{head}$ for every token $i$
\If{ $\forall i$, $S_i^\text{head} \leq \theta_{\text{prune}}(\alpha \cdot R)$ }
    \State \textbf{Token Pruning}: Obtain $\mathbf{\hat{V}}^l_{\text{prune}}$ via \cref{eq:token_pruning}
\EndIf
\If{ $\forall i$, $\theta_{\text{merge}}(R) \leq S_i^\text{head} \leq \theta_{\text{prune}}(\alpha \cdot R)$ }
    \State \textbf{Token Merging}: Obtain $\mathbf{\hat{V}}^l_{\text{merged}}$ via \cref{eq:token_merged}
\EndIf\\
\Return $\mathbf{\hat{V}}^{l}$, which is composed of $\mathbf{v}^l_{\operatorname{cls}}$, $\mathbf{\hat{V}}^l_{\text{prune}}$ (excluding merged tokens), and $\mathbf{\hat{V}}^l_{\text{merged}}$
\end{algorithmic}
\end{algorithm}

\begin{figure*}[t]
    \centering
    \includegraphics[width=0.9\linewidth]{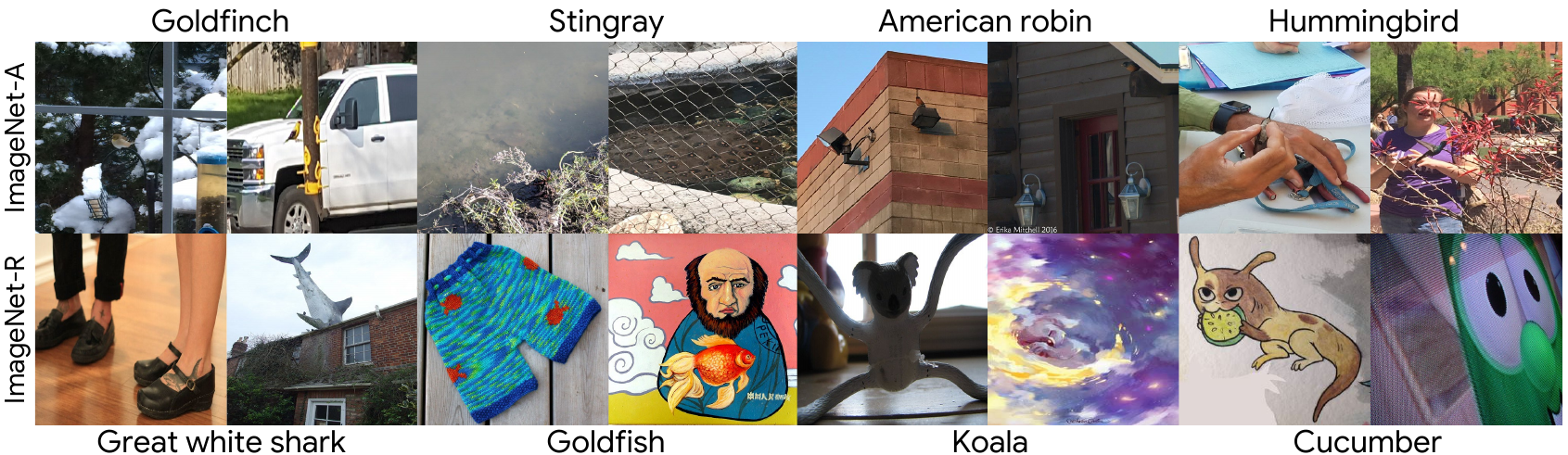}
    \caption{\textbf{Sample data from the OOD benchmark.} The samples from the same class exhibit significant diversity. For instance, in the ImageNet-R dataset, one image of a great white shark is dominated by shoes and human legs, while another is on top of a building, showing extreme variability. }\label{fig:ood_sample}
\end{figure*}
\textbf{\cref{alg:ours}} outlines the process for performing token pruning and merging at layer $l$ in a ViT-based CLIP model. We first obtain the averaged domain anchor tokens $\mathbf{A}^{l-1}_{c^*}$ by the \texttt{<cls>} tokens saved in the reservoir $\mathfrak{R}$. Token condensation is then conducted given the domain anchor token. Specifically, we conduct token pruning by relative ranking positions of token $i$ across multiple attention heads. Then, coreset selection is used for token merging. Finally, we concatenate the \texttt{<cls>} token $\mathbf{v}^l_{\operatorname{cls}}$ with the retained tokens as the input for the next layer, where the original $N+1$ tokens are shrunk to $(R \cdot N) +1$, thereby reducing the computational cost.

\subsection{Quantitative Study}\label{sec:q_stu}
\begin{figure*}[t]
    \centering
    \begin{subfigure}[t]{0.9\linewidth}
        \centering
        \includegraphics[width=\linewidth]{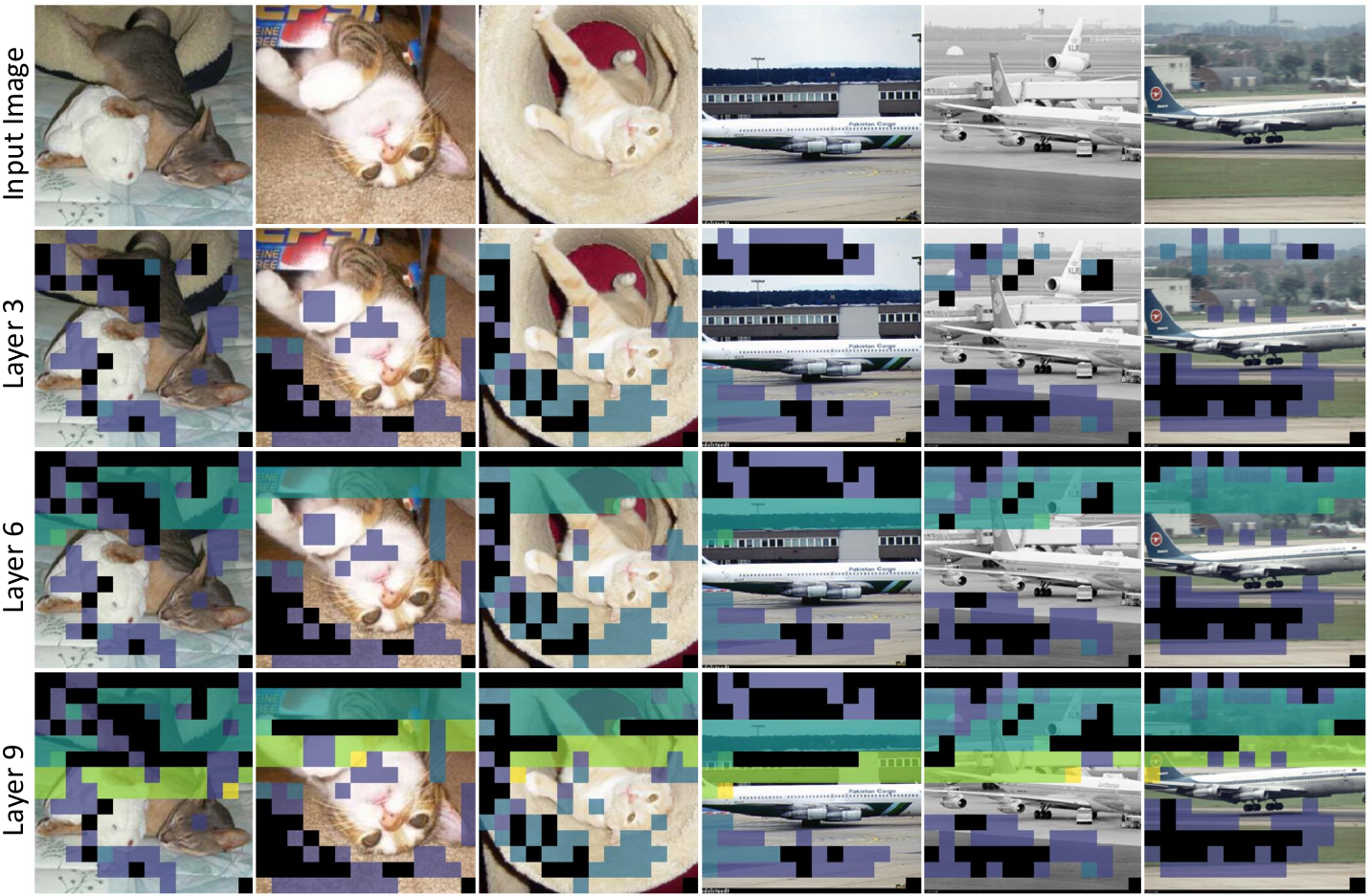}
    \end{subfigure}%
    \\
    \begin{subfigure}[t]{0.9\linewidth}
        \centering
        \includegraphics[width=\linewidth]{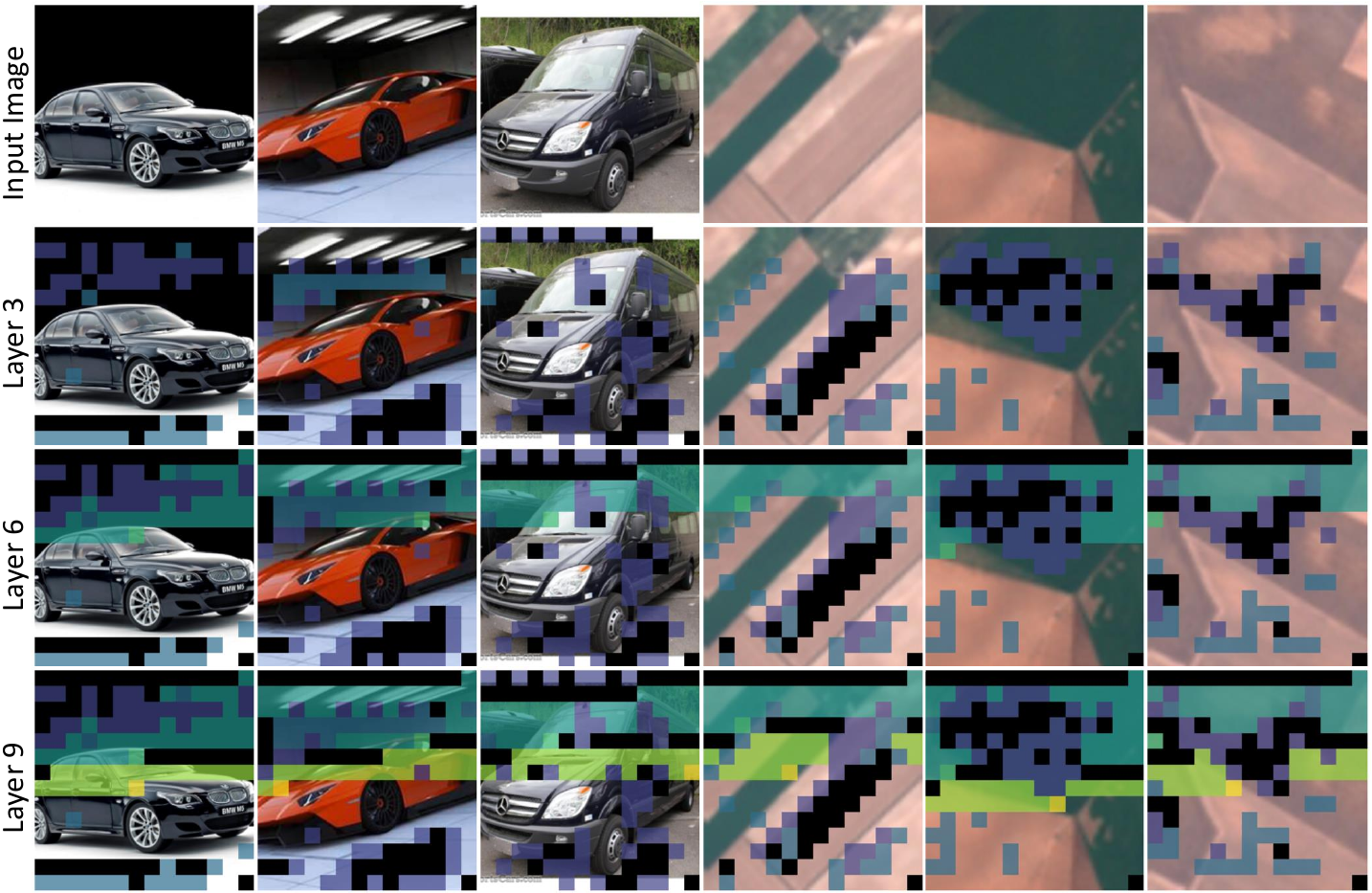}
    \end{subfigure}
    \caption{Visualization of our proposed token condensation with $R=0.7$. Pruned tokens are masked in black, while different colors represent distinct merging clusters.}
    \label{fig:token_viz}
\end{figure*}

We visualize the token condensation masks at layer 3, layer 6, and layer 9, and compare them with the original image across multiple datasets, as shown in \cref{fig:token_viz}.  As the layers go deeper, we observe that class-irrelevant patches are gradually pruned, as indicated by the black mask. TCA also merges class-ambiguous patches, such as fur in cat images, and ground and sky in aircraft and car images. All similar tokens are merged into a single token using our proposed coreset selection strategy. After token condensation, the sample features retain only discriminative information, thereby bridging the gap between visual and text features, and mitigating the distribution shift between pretrained data and unseen datasets.

\subsection{Discussion on TCA's Generalizability} \label{sec.general}
TCA is designed for VLMs such as CLIP, SigLIP, and SigLIP v2, requiring only minor modifications. These models share a key characteristic: they compute cosine similarity between modalities for zero-shot image classification. For CLIP, we use the \texttt{<cls>} token as a guiding indicator throughout the method. In contrast, for the SigLIP series, we take the average over attention weights since their architecture does not include a visual \texttt{<cls>} token. The way we determine the domain anchor token and perform token condensation is inherently tied to how each VLM extracts visual features for alignment. We acknowledge that TCA may not directly apply to models like LLaVA \cite{DBLP:conf/nips/LiuLWL23a}, as they are not designed for cross-modal alignment but rather for text generation, dictated by their architectural constraints. While this limits direct applicability, it does not diminish TCA's effectiveness in its intended scope. Adapting it to such models would likely require a fundamental architectural redesign.

\subsection{Discussion on the Limitation of TCA}\label{sec:limitation}
In this section, we discuss the potential limitations of our proposed TCA. Due to the training-free nature of the approach, it is challenging to mitigate the performance gap when the testing domain diverges significantly from the training domain. As observed in the out-of-distribution (OOD) samples shown in \cref{fig:ood_sample}, the ground truth object is not always centrally located, and larger class-irrelevant objects (\textit{e.g.,} humans or shoes) can sometimes dominate the prediction. This issue is particularly prominent in CLIP models, where text features for all classes are predefined. When the dominant object is included in the label set, accurately directing visual features to the correct class without additional training becomes difficult. Moreover, the diversity of OOD samples introduces further complexity, especially in the absence of data augmentation. These observations raise important questions for future research: (1) How can we quantify the capacity to mitigate domain shift effectively? (2) What lightweight solutions can be developed for backpropagation and network updates to facilitate test-time adaptation? We leave these questions for future work.